\documentclass[reqno]{amsart}
\DeclareRobustCommand{\SkipTocEntry}[5]{}

\usepackage{styleA}
\usepackage{orcidlink}

\usepackage[margin=3cm]{geometry}

\makeatletter
\def\blfootnote{\gdef\@thefnmark{}\@footnotetext}
\makeatother

\title[]{Detection of Anomalous Vehicular Traffic and Sensor Failures Using Data Clustering Techniques}

\author[D.\ Moretti, E.\ Onofri, and E.\ Cristiani]{}

\date{}

\begin{document}

\blfootnote{$^*$Corresponding author \href{mailto:davide.moretti@cnr.it}{\Envelope}
\texttt{davide.moretti@cnr.it}}

\maketitle

\vspace{-1em}

\begin{center}
    \begin{minipage}{.89\linewidth}\centering
        \textsc{Davide Moretti}$^{1, 2, \orcidlink{0009-0001-3503-3879}, *}$
        \hspace{2em}
        \textsc{Elia Onofri}$^{1, \orcidlink{0000-0001-8391-2563}}$
        \hspace{2em}
        \textsc{Emiliano Cristiani}$^{1, \orcidlink{0000-0002-7015-2371}}$
        \\[1.5em]
        {\footnotesize
            $^1$
            Istituto per le Applicazioni del Calcolo (IAC),
            National Research Council of Italy (CNR)
            \\\smallskip
            $^2$
            Dipartimento di Matematica e Fisica, University of Roma Tre, Italy
        }
    \end{minipage}
\end{center}

\medskip
\thispagestyle{empty}

\begin{abstract}
The increasing availability of traffic data from sensor networks has created new opportunities for understanding vehicular dynamics and identifying anomalies. In this study, we employ clustering techniques to analyse traffic flow data with the dual objective of uncovering meaningful traffic patterns and detecting anomalies, including sensor failures and irregular congestion events.

\medskip\noindent
We explore multiple clustering approaches, \ie partitioning and hierarchical methods, combined with various time-series representations and similarity measures. Our methodology is applied to real-world data from highway sensors, enabling us to assess the impact of different clustering frameworks on traffic pattern recognition. We also introduce a clustering-driven anomaly detection methodology that identifies deviations from expected traffic behaviour based on distance-based anomaly scores.

\medskip\noindent
Results indicate that hierarchical clustering with symbolic representations provides robust segmentation of traffic patterns, while partitioning methods such as $k$-means and fuzzy $c$-means yield meaningful results when paired with Dynamic Time Warping. The proposed anomaly detection strategy successfully identifies sensor malfunctions and abnormal traffic conditions with minimal false positives, demonstrating its practical utility for real-time monitoring.

\medskip\noindent
Real-world vehicular traffic data are provided by Autostrade Alto Adriatico S.p.A.    

\bigskip


\medskip

\noindent\textbf{MSC-2020.}
\textbf{90B20}, 
62M10

\end{abstract}

%\begin{multicols}{2}
    \vspace{-1.3em}
    \tableofcontents
    \vspace{-2em}
%\end{multicols}

\newpage

\section{Introduction}\label{sec:intro}

The rapid proliferation of sensors in modern technology has led to an unprecedented surge in data generation. From industrial monitoring systems to wearable devices and smart cities, sensors continuously collect vast amounts of information. These data streams are inherently temporal and often unstructured and unlabelled, requiring specialised techniques to extract meaningful insights.
The intricate structure of time series often conceals valuable patterns and relationships from superficial analysis, hence requiring advanced data mining techniques.
Consequently, time series data mining has found applications in multiple disciplines for tasks such as
anomaly detection,
motif and pattern discovery,
indexing,
clustering,
classification,
visualisation,
trend analysis,
and forecasting~\cite{mitsa2010temporal}.

Among these tasks, \textit{clustering} plays a crucial role in uncovering latent structures within unlabelled temporal datasets. It involves partitioning data into homogeneous groups based on a similarity measure that quantifies how ``close'' the data objects are in a given space. The choice of a suitable (set of) distance function(s) is often application-dependent, and selecting the appropriate one, along with a representation space, is a challenge in itself.
Furthermore, datasets suitable for data mining techniques are usually large and high-dimensional, demanding significant computational resources.

A notable example of large-scale time series data generation comes from traffic sensors deployed on highways. Such sensors continuously monitor and record vehicle flow passing beneath them, capturing key attributes such as vehicle counting and corresponding speeds as well as grouping vehicles by class (\eg light and heavy vehicles).
These pieces of information are crucial for traffic management, congestion prediction, and road safety analysis, and are typically used in traffic model calibration and validation.
However, the data collected by these sensors are inherently temporal, vast in size, high-dimensional, and often unlabelled, making it challenging to analyse using traditional methods.
For instance, even distinguishing between an empty road and a fully congested one --both scenarios in which no vehicles pass the sensor-- presents an interesting challenge that heavily relies on the temporal nature of data~\cite{Briani2024}.

In this context, clustering techniques can help uncover patterns in traffic behaviour, such as recurring congestion events, seasonal variations, or anomalous fluctuations that might indicate accidents or roadwork.
While extensive research has been devoted to applying clustering in traffic analysis, little work has been done on anomaly detection; most studies assume that sensors operate flawlessly and do not account for the possibility of sensor failures in their models. However, real-world traffic monitoring systems are prone to a wide range of sensor malfunctions that can severely impact data reliability. These failures manifest in various forms, some of which are not immediately evident. Common issues include missing data, incorrect temporal aggregation (\eg, misaligned timestamps), abnormally large spikes in recorded values (caused \eg by sudden fluctuation on the sensor power line), or even data streams that remain constant over time despite expected variations. Such anomalies can introduce significant biases in traffic models and, if undetected, lead to inaccurate conclusions in downstream analyses.

Detecting sensor failures is particularly challenging because erroneous data may still appear plausible when considered on its own. For instance, a sensor might report a realistic number of passing vehicles or average speeds, yet these values could be inconsistent w.r.t.\ historical trends or be in conflict with data from neighbouring sensors. Given the diversity of potential failure modes, it is unreasonable to assume that all possible faults can be identified \textit{a priori}. Instead, automated approaches capable of dynamically detecting anomalies are needed. 
By revealing unexpected deviations in traffic patterns, clustering techniques offer a promising solution.

The use of clustering for anomaly detection, however, comes with various challenges.
Clustering results still require expert validation before being operationalised, as repeated anomalies can potentially accumulate into a distinct, well-formed cluster.
This phenomenon is particularly problematic in clustering methods that require a predefined number of clusters; if a method is expected to identify $k$ typical clusters but instead detects $\hat{k}$ anomalous clusters, then the number of discernible normal behaviours is effectively reduced to $k - \hat{k}$.
Consequently, some meaningful traffic patterns may go undetected.
This issue becomes even more critical when anomalies evolve into persistent trends, making it difficult to distinguish between transient faults and fundamental shifts in sensor behaviour (\eg a sensor whose miscalibration goes unnoticed).
Addressing these challenges requires ad-hoc strategies to ensure that clustering techniques remain robust in dynamic environments where anomalies might transition into new steady-state behaviours.

\subsection{Relevant literature}

\paragraph{Clustering}

Clustering is a fundamental unsupervised learning technique that partitions data into groups, or clusters, based on some notion of similarity measure.
Classical clustering algorithms can be broadly categorised into partition-based, hierarchical, density-based, and model-based approaches~\cite{Aghabozorgi2015, Warren05}.
Among partition-based methods, $k$-means~\cite{macqueen1967some} and fuzzy $c$-means~\cite{bezdek1973cluster} remain two of the most widely used approaches due to their efficiency and simplicity, particularly when applied to time series data~\cite{Aghabozorgi2015}.
However, they suffer from limitations such as assuming spherical clusters (for some space embedding) and requiring the number of clusters to be specified in advance.
Hierarchical clustering methods, like, \eg, agglomerative clustering, allow for a flexible exploration of cluster granularity without requiring a predefined number of clusters~\cite{johnson1967hierarchical}.
Density-based approaches, such as DBSCAN~\cite{ester1996density}, are particularly useful in detecting clusters of arbitrary shape and are well-suited for handling noise and outliers.
On the other hand, model-based techniques, such as Gaussian Mixture Models (GMMs)~\cite{bishop2006pattern}, assume an underlying probabilistic structure and provide soft clustering assignments, making them particularly effective when clusters exhibit overlapping distributions.

Despite significant advancements in clustering methodologies, their effectiveness heavily depends on selecting an appropriate similarity measure, which must be coupled with a suitable embedding space to represent the data.
In traditional clustering applications where data points reside in Euclidean spaces, the $L_2$ norm provides an intuitive, robust, and reliable similarity measure.
However, in more complex settings, particularly with time series data, defining meaningful distances becomes a significant challenge.
Time series are inherently sequential, high-dimensional, and prone to temporal distortions, necessitating the development of specialised similarity measures and embedding techniques~\cite{Warren05}.

\paragraph{Time series clustering}

Existing approaches for time series clustering can be divided into three main categories:
\begin{description}
  \item[Whole time series clustering] which treats each time series as a single entity and applies classical similarity measures such as the Euclidean distance (easier yet unable to capture temporal distorsions) or more advanced metrics, like, \eg, the Dynamic Time Warping (DTW), originally introduced as a speech pattern recognition technique in~\cite{sakoe1978dynamic} and later applied in many other fields \cite{berndt1994using}.

  \item[Subsequence clustering] which segments time series into smaller subsequences before clustering. This approach is useful for detecting repeated patterns, such as motifs in physiological or financial data, but can suffer from spurious patterns due to overlap~\cite{keogh2003needles}.

  \item[Feature-based clustering] which transforms time series into a lower dimensional space using statistical descriptors~\cite{fulcher2013highly}, wavelet coefficients~\cite{Aghabozorgi2015}, or shapelet-based representations~\cite{ye2009time}, improving efficiency and interpretability while reducing computational complexity.
\end{description}

The latter approach is particularly useful when dealing with large-scale datasets, where direct similarity evaluations are computationally prohibitive, as it happens in our case study.
Amongst the simplest feature-based techniques, Piecewise Aggregate Approximation (PAA)~\cite{keogh2001dimensionality} provides an efficient way to reduce dimensionality by summarising segments of time series with average values. This technique has been successfully applied in numerous domains and later combined within Piecewise Dynamic Time Warping (PDTW)~\cite{keogh2000scaling}, an adaptation of DTW designed for compressed representations~\cite{chu2002iterative}.

Building on these ideas, symbolic representations have gained traction for their efficiency in handling large-scale datasets.
Symbolic Aggregate approXimation (SAX)~\cite{lin2003symbolic} discretises continuous values into symbolic categories based on a Gaussian-spaced grid, naturally coupled with $\mindist$, a tailored distance metric confronting symbols on the underlying grid~\cite{lin2007experiencing}.
A large amount of work was performed in this direction, introducing numerous variants including the Extended SAX (ESAX)~\cite{lkhagva}, which retains also information about minima and maxima of the interval, the Indexable SAX (iSAX)~\cite{shieh2008sax}, particularly suited for terabyte-sized time series, and the SAX-trends \cite{sun2014improvement}, incorporating trend information to enhance clustering accuracy.

\paragraph{Evaluating clustering performance}

When it comes to assessing clustering kindness, many different methodologies and indicators can be employed.
With some labelled examples provided, classical solutions like precision, accuracy, and recall can be used, potentially paired with the analysis of the corresponding ROC curves.
However, when no information is available on correct data classification, a solution that takes into account intra- and inter-cluster discrepancies must be used.
Amongst the mostly used, we recall the \emph{Silhouette index}, originally introduced in 1987 in~\cite{rousseeuw1987silhouettes} as an indicator to distinguish between well- and badly-separated clusters, later adopted also for non-Euclidean metrics, see \eg~\cite{staples2023reproducible}.
Also worth of mention is the \textit{Partition Coefficient and Exponential Separation} (PCAES) which, originally introduced in \cite{wu2005cluster}, is specifically useful in the context of fuzzy clustering, \ie when point are not assigned to a specific cluster only, but rather the method provides the probability of a given point being in any specific cluster.
For a general treatise on the usage and comparison of cluster validity indices, we refer the interested reader to \cite{arbelaitz2013extensive}.

Additionally, hierarchical clustering methods provide insights into clustering robustness by analysing the stability of significant clusters across different levels. For instance, in \cite{Onofri23}, the author explored how monitoring the number of $p$-significant clusters can reveal patterns in data segmentation.

\paragraph{Anomaly detection}

Anomaly detection is a fundamental task in data analysis, aiming to identify unusual patterns, rare events, or deviations from expected behaviour~\cite{Blazquez21}.

Among the various approaches to anomaly detection, clustering-based methods provide a natural way to distinguish between normal and abnormal instances.
Here, clusters represent groups of typical patterns, while points deviating significantly from their nearest clusters are flagged as anomalies.
The distance from the closest cluster centroid is commonly used as an anomaly score, often referred to as \textit{reconstruction error}~\cite{6608627}.
This method is particularly effective in unsupervised settings, where anomalies are unknown a priori and must be inferred from the structure of the data.

Beyond centroid-based techniques, alternative strategies have been developed to improve robustness, including (i) Density-Based Anomaly Detection, (ii) Reconstruction-Based Approaches, and (iii) Time Series Anomaly Detection.
Within the latter category, methods such as Local Outlier Factor (LOF)~\cite{Breunig2000} and DBSCAN-based outlier scoring~\cite{ester1996density} identify anomalies by analysing local density variations.
Conversely, techniques such as autoencoders~\cite{Chen2018} and Principal Component Analysis (PCA)-based reconstruction errors~\cite{Shyu2005} have been used to detect anomalies as deviations from learned representations.
Finally, in particular for sequential data, distance measures such as Dynamic Time Warping (DTW) anomaly scores~\cite{Diab2019} or forecasting residuals from Long Short-Term Memory (LSTM) networks~\cite{Markovic2023} highlight unexpected temporal variations.

In what follows, we adopt a clustering-driven anomaly detection approach, using distance-based anomaly scoring to measure deviations from expected patterns.

\paragraph{Traffic analysis}

Managing, understanding, and predicting traffic is a fundamental aspect of transportation planning, infrastructure development, and smart city initiatives~\cite{Ganapathy2021-zw}.
Traffic data are inherently temporal, comprising information such as vehicle count, speed, congestion levels, and travel times, often collected across multiple locations in a road network (spatial component).
Extracting meaningful insights from these data is crucial for optimising traffic flow, reducing congestion, improving road safety, and supporting intelligent transportation systems (ITS)~\cite{Banihosseini2024-bk}.

In recent years, the increasing availability of high-resolution traffic data from sensors, IoT devices, and connected vehicles has led to a paradigm shift in traffic analysis. In~\cite{MedinaSalgado2022}, the authors show how traditional methods, which adopt aggregated and low-frequency data, are progressively being replaced by real-time monitoring and forecasting techniques that leverage machine learning and artificial intelligence.

While real-time forecasting focuses on short-term traffic dynamics, offline approaches remain crucial for long-term traffic pattern analysis, strategic infrastructure planning, and anomaly detection.
Early approaches include, for example, \cite{Weijermars2005-tj}, where the authors analysed historical highway traffic flow data using clustering to distinguish common patterns repeating across different locations due to daily and seasonal features.
In particular, the approach of analysing flux data within the various clustering frameworks also received considerable attention within the recent literature.
It is the case, \eg, of \cite{Asadi2019} where sub-sequences of traffic data are clustered performing an embedding to reduce the dimensionality and where DTW is used to compare the features vectors. A similarity heatmap is then created between sensors to reconstruct spacial similarity from the temporal clusters. In a similar fashion,~\cite{Toshniwal2020} tries to capture the underlying traffic pattern and phenomena by spatially and temporally clustering flux data gathered in Aarthus (Denmark) while trying to correlate different behaviours with different street types.
Also worth attention is the work~\cite{LI2022266}, where an analysis of urban traffic data is performed, offering a comparison between different possible clustering approaches on the same dataset. Here, both internal validation indices and macroscopic pieces of information about roads, vehicles and people are applied to evaluate and validate the results. 

However, despite this progression, time-series clustering application in identifying traffic anomalies --such as sensor malfunctions or unexpected disruptions-- has received less attention.
Sensor failures in traffic monitoring systems can lead to inaccurate data collection, ultimately affecting downstream analytics and decision-making.
Developing automated, clustering-based approaches for detecting anomalous traffic sensor behaviour could provide a robust, data-driven solution for ensuring data integrity in large-scale traffic monitoring networks.

\subsection{Contributions}

In the present work, we present a comprehensive study on clustering-based traffic analysis, focusing on the identification of typical traffic patterns and anomaly detection using real-world highway sensor data.
The key contributions of this study are as follows:
\begin{description}
    \item[Comparison of clustering techniques] We conduct an extensive evaluation of various clustering methodologies, including partitioning ($k$-means, fuzzy $c$-means) and hierarchical approaches, applied to time-series traffic data. Our analysis highlights the impact of different time-series representations and similarity measures on clustering effectiveness.
    
	\item[Clustering-driven anomaly detection] We propose a novel approach that integrates clustering results with anomaly detection mechanisms. By analysing deviations from cluster centroids, our method effectively identifies irregular traffic patterns, with a particular focus on sensor malfunctions and traffic flux anomalies.
    
	\item[Multivariate time-series analysis] We extend the clustering framework to a multivariate setting, incorporating data from multiple sensor sources and multiple vehicle classes aggregation. This approach improves the detection of complex anomalies that are not apparent in univariate analyses.
    
  \item[Real-world application] Our methodologies are applied to a dataset of traffic data collected by highway sensors dispatched in northern Italy, prompting insights into real traffic conditions.
  We also develop an automated pipe\-line for traffic pattern analysis and anomaly detection, providing a practical tool for practitioners to monitor and ensure the integrity of traffic data.
\end{description}

\subsection{Paper organisation}
The rest of the paper is organised as follows.
In Section~\ref{sec:dataset}, we introduce the dataset used in the study, including data sources, preprocessing steps, and initial univariate (Section~\ref{ssec:discussing-univariate}) as well as multivariate (Section~\ref{ssec:multivariate}) analyses.
In Section~\ref{sec:methods}, we present the clustering methodologies, time-series representations, similarity measures, and evaluation criteria adopted in our study (Section~\ref{ssec:clustering-method}). 
We also describe the novel anomaly detection framework based on the clustering results (Section~\ref{ssec:Anomaly Detection}).
In Section~\ref{sec:results}, we discuss the findings of our clustering experiments (Section~\ref{ssec:clustering-results}), compare different methodologies, and evaluate the performance of our anomaly detection system (Section~\ref{ssec:anomaly-results}). Representative case studies of detected anomalies are also discussed.
Finally, in Section~\ref{sec:conclusions}, we summarise the key findings of the study, discuss its implications, and outline potential directions for future research.
The paper concludes with Appendix~\ref{app:novel-averaging}, where we discuss a previously unpublished approach to averaging symbolic series.

\section{Discussing the real-world data}\label{sec:dataset}

The dataset used in this paper is provided by Autostrade Alto Adriatico S.p.A. (previously Autovie Venete S.p.A.), which manages various highway segments in northern Italy, namely the A4 Venezia\---Trieste, A23 Palmanova\---Udine Sud, A28 Portogru\-aro--Conegliano, A34 Villesse\---Gorizia and part of the A57 Tangenziale di Mestre, for a total road length of roughly 234 km.

The traffic flow in the aforementioned area is monitored via a large number of cameras, mobile sensors, and fixed sensors. 
Fixed sensors are gathered into \textit{stations}, which are placed along the highway in fixed positions. 
Each station is associated with a single traffic direction, and the distance between them is not constant, ranging from 2 to 20 km. Sensors in different stations can be based on different technologies, but they perform in a standardized way:
each sensor in each station monitors \textit{a single lane}, providing traffic data every minute, which includes flux (number of vehicles passed under the sensor), average velocity, and occupancy rate. 
The relative error on counting and velocity claimed by the manufacturer is of $\pm3\%$.

In this paper, we only consider flux data, which are further divided on the basis of the class of vehicle, following the German TLS 5+1 class standard\footnote{\url{https://www.bast.de/DE/Publikationen/Regelwerke/Verkehrstechnik/Unterseiten/V5-tls.html}}.
We aggregate classes 1 and 2 in a new macro-class named \textit{light vehicles} and classes 3, 4, and 5 in another macro-class named \textit{heavy vehicles}.
An aggregate class 0 is also provided, obtained by summing the total vehicular flux in the time interval and performing a weighted average for the speed and occupancy rates. 

\begin{figure*}[t]
\centering
\begin{subfigure}{0.5\textwidth}
  \centering
  \includegraphics[width=.99\linewidth]{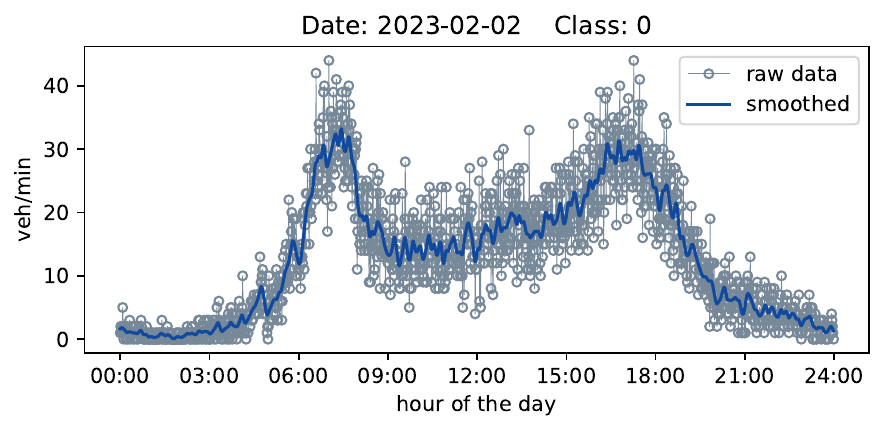}
  \label{fig:exn1}
\end{subfigure}%
\begin{subfigure}{0.5\textwidth}
  \centering
  \includegraphics[width=.99\linewidth]{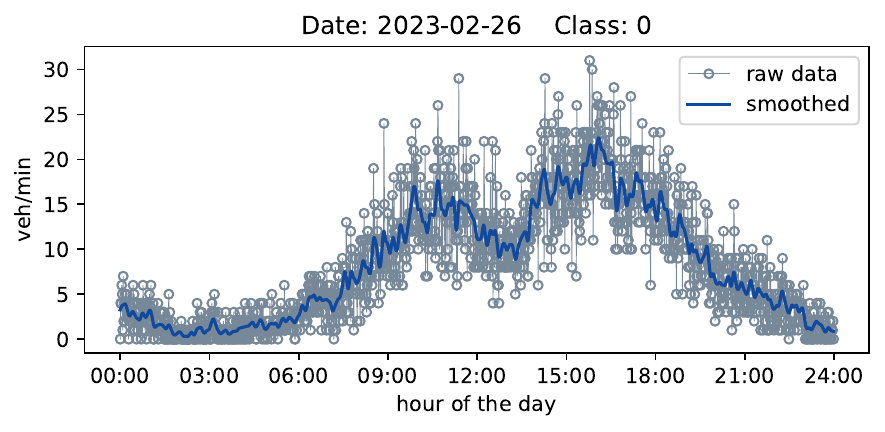}
  \label{fig:exn2}
\end{subfigure}
\caption{
    Examples of normal single-day time series.
    The characteristic double-hump behaviour of flux data can be seen in different time frames. 
    Do notice that, for the sake of readability, $y$-axes are not homogeneous.
}
\label{fig:exn}
\end{figure*}
\begin{figure*}[t]
\centering
\begin{subfigure}{.49\textwidth}
  \centering
  \includegraphics[width=.99\linewidth]{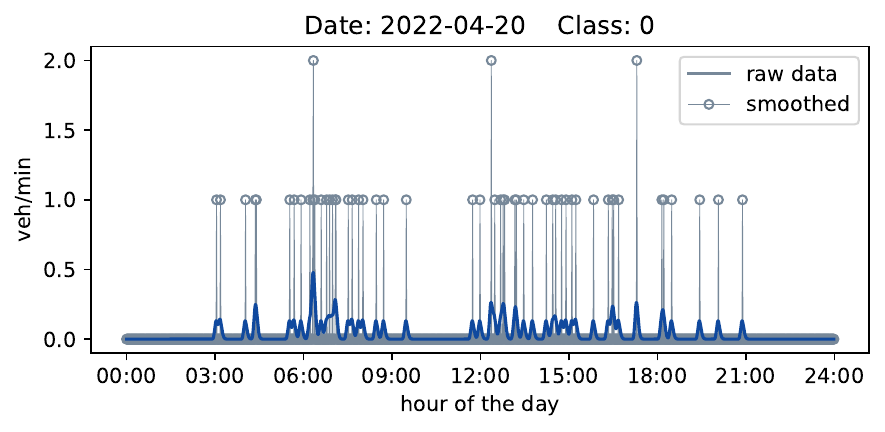}
  \label{fig:exa11}
\end{subfigure}%
\begin{subfigure}{.49\textwidth}
  \centering
  \includegraphics[width=.99\linewidth]{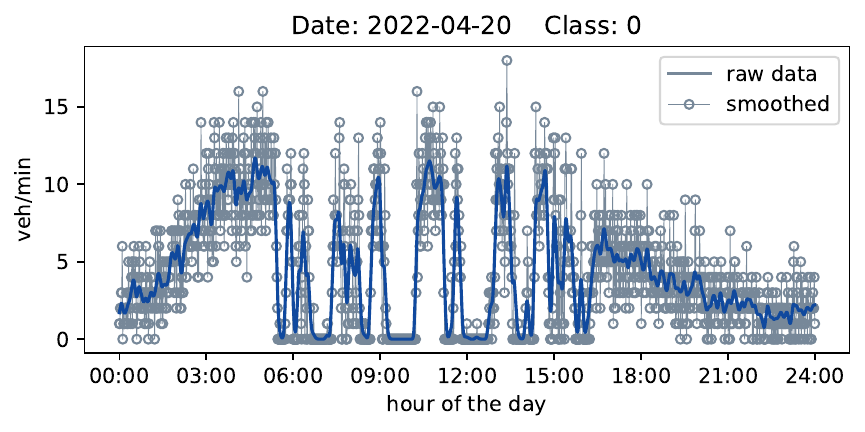}
  \label{fig:exa12}
\end{subfigure}
\begin{subfigure}{.49\textwidth}
  \centering
  \includegraphics[width=.99\linewidth]{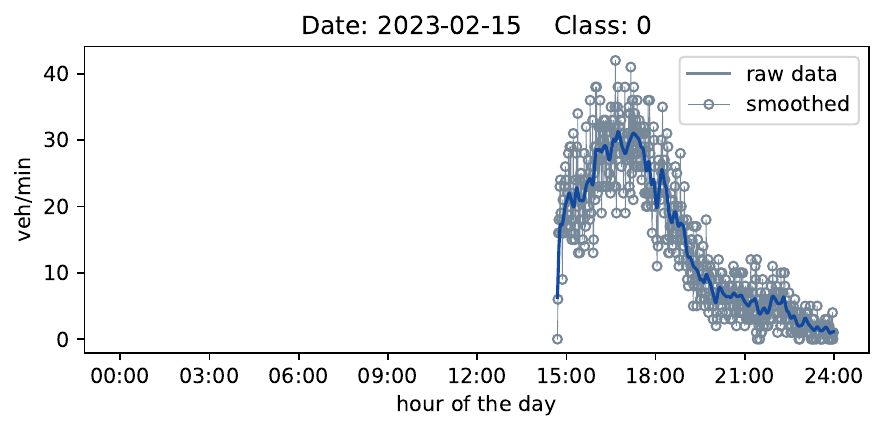}
  \label{fig:exa21}
\end{subfigure}%
\begin{subfigure}{.49\textwidth}
  \centering
  \includegraphics[width=.99\linewidth]{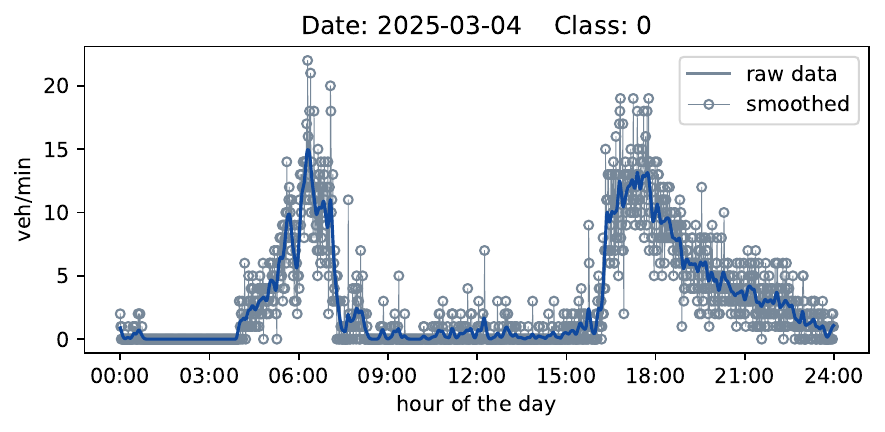}
  \label{fig:exa22}
\end{subfigure}
\caption{Examples of anomalous one-day behaviour. On the left, probable sensor failures that are automatically removed from the dataset. On the right, probable traffic anomalies, which are harder to detect automatically. Do notice that, for the sake of readability, $y$-axes are not homogeneous.}
\label{fig:exa}
\end{figure*}

\subsection{Univariate analysis}\label{ssec:discussing-univariate}
For the basic analysis, we focus on aggregated flux data in class 0. 
Our dataset is composed of all the daily time series of flux data measured by every sensor and spanning the entire year of 2022. 
The ``atomic datum'' of our analysis is, therefore, a time series of length 1440 (one flux datum for every minute of the day).
Figure~\ref{fig:exn} shows visual examples of ``normal'' data, \ie data expected for a usual day with no sensor anomalies, no relevant accidents, and no traffic congestion; a smoothed version of the data obtained via convolution with a Gaussian kernel is also depicted in the figure for better readability.

\medskip

Along with regular data, it is important to note that daily time series might present different anomalies, some of which are worth further investigation during the dataset construction and whose removal would yield clearer results.

One significant example is represented by the presence of an abnormal number of minutes in which flux zero is recorded.
Such a situation, see Figure~\ref{fig:exa}(top-left), is typically hard to explain without contextual information, like \eg the presence of a construction site or some kind of lane closure; hence, we decided to remove time series presenting a total flux not exceeding a given small threshold.

A different example is given by a time series with missing flux data.
This outcome might be related to a sensor failure, a connection failure, or a database storage failure. 
Although in some cases realistic values might be inferred, such a procedure is not completely trustworthy and might not be applicable if gaps are long in time, see \eg Figure~\ref{fig:exa}(bottom-left).
Hence, we decided to remove any time series with any missing data to form the cleaned dataset we later used for clustering analysis.

Finally, it is worth noting that traffic anomalies might also appear.
However, differently from the previous two cases, no simple method for automatic detection can be mounted.
As an example, do consider that a flux reading of zero could be transmitted by the sensor due to two different opposing situations: (i) the road is empty and no vehicle passed under the sensor or (ii) vehicles are completely stopped and no one pass through the sensor due to a full congestion (queue); similar outcomes can be derived for contexts with low fluxes and discriminating the two cases is an interesting problem \textit{per se}, also when velocities are considered, as discussed in~\cite{Briani2024}.
Figure~\ref{fig:exa}(right) depicts two different traffic anomalies where both these situations can be spotted: at night, we have some null flux data due to the empty road, while during the day, we have some null flux data caused by full congestions.
Due to the intrinsic difficulties in detecting them, such anomalies are amongst the most interesting ones in our scenario and, although we removed some of them via heuristic and manual analysis, a consistent number is still present in the cleaned dataset.

A final cleaned dataset comprising 31,125 daily time series forms the base of our univariate analysis.

\subsection{Multivariate analysis}\label{ssec:multivariate}
\begin{figure*}
    \centering
    
    \includegraphics[height=3cm]{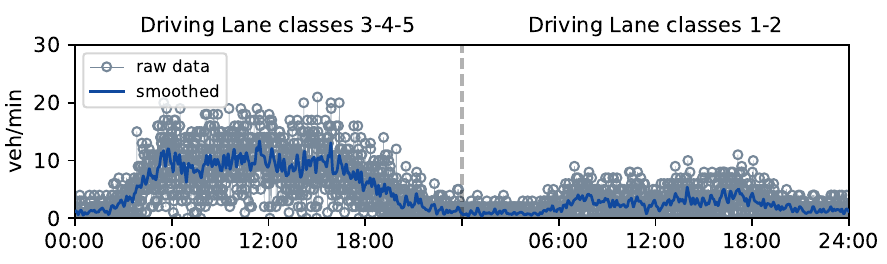}\\
    \includegraphics[height=3cm]{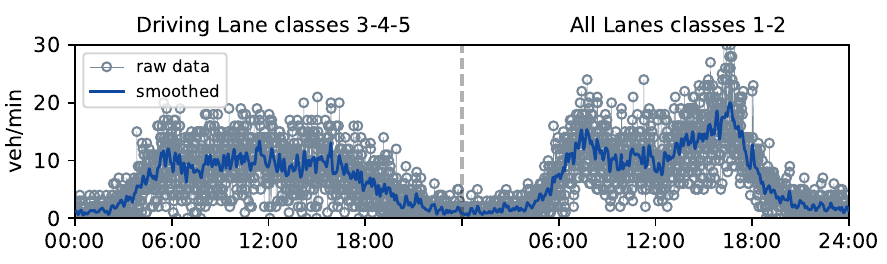}\\
    \includegraphics[height=3cm]{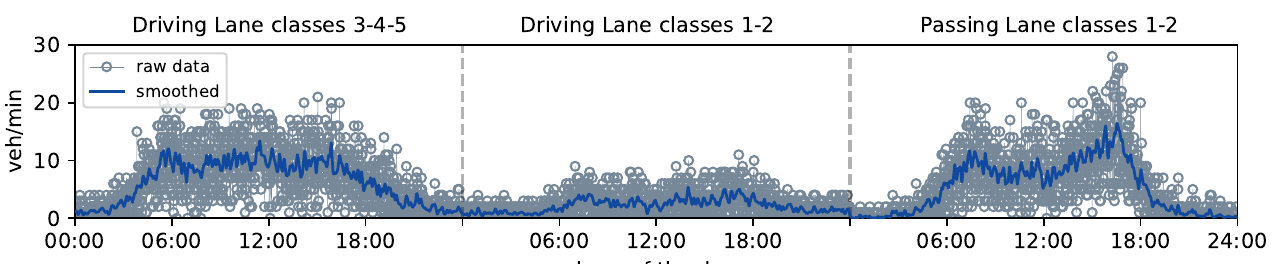}
    
    \caption{
        Example of normal single-day multivariate time series for the three types of analysis introduced in Table~\ref{table:multivariate} (T1 -- T3 from top to bottom).
        Each box within a panel corresponds to one variable
    }
    \label{fig:multivariate-exn}
\end{figure*}

For a more advanced and detailed analysis, we also consider finer data coming from multiple sensors simultaneously, yet belonging to the same station and the same day.
The ``atomic datum'' is, therefore, the combination of two or more daily time series derived from sensor data after a partial lane and/or class aggregation.
Amongst the various attempts we explored in our analysis, we here report the three most promising types of multivariate data aggregations, a summary of which can be found in Table~\ref{table:multivariate}.
In particular, we considered heavy vehicles on the sole driving (slow) lane as \textit{Variable A}: here, it is important to recall that heavy vehicles are not theoretically allowed on the passing (fast) lane.
Then, we paired \textit{Variable A} with (T1) light vehicles in the same lane, (T2) light vehicles in the same station, and (T3) light vehicles divided between the driving and passing lanes.
Figure~\ref{fig:multivariate-exn} depicts a typical behaviour of each approach.

\begin{table*}[t]
    \centering
    \caption{
        Different approaches for the multivariate analysis.
    }
    \setlength{\tabcolsep}{4pt}
    \renewcommand{\arraystretch}{1.2}
    \begin{tabular}{C{1.5cm}C{3.5cm}C{3.5cm}C{3.5cm}}
        \toprule
        \textbf{Type} & \textbf{Variable A} & \textbf{Variable B} & \textbf{Variable C} \\
        \midrule
        \multirow{2}{*}{T1} & driving lane & driving lane & \multirow{2}{*}{--} \\ 
        & heavy vehicles & light vehicles & \\
        \midrule
        \multirow{2}{*}{T2} & driving lane & all lanes & \multirow{2}{*}{--} \\ 
        & heavy vehicles & light vehicles & \\
        \midrule
        \multirow{2}{*}{T3} & driving lane & driving lane & passing lane \\ 
        & heavy vehicles & light vehicles & light vehicles \\
        \bottomrule
    \end{tabular}
    \label{table:multivariate}
\end{table*}

For this analysis, we have considered a small dataset made of 158 days from a single station.
Despite the small size of the dataset, it allows us to provide a proof of concept of how a higher level of detail actually improves traffic data analysis.

\section{Methods}\label{sec:methods}
In order to provide a clear understanding of the proposed analysis, in this section we present an overview of the implemented methods, which are later compared.
The confident reader can safely skip to Section~\ref{ssec:Anomaly Detection}, where our novel approach to anomaly detection is presented.

\subsection{Clustering}\label{ssec:clustering-method}

In what follows, we focus specifically on PAA, SAX, ESAX, MINDIST, DTW, $k$-means\plusplus{}, fuzzy $c$-means\plusplus{}, agglomerative hierarchical clustering, and clustering result evaluation. 

\paragraph{Time series representation: PAA}\label{Piecewise Aggregate Approximation PAA}
It is one of the simplest time-series representations, capable of dimensionality reduction whilst still matching the original shape of the series. A time series $s=\{x_1\dots x_n\} \in \mathcal{S}$ of length $n$ can be represented in a $w$-dimensional space $\mathcal{S'}$ (with $w\leq n$) as a vector $\overline{s} = \{\overline{x}_1\dots \overline{x}_w\}$ with the $i$-th element of the sequence being
\begin{equation}\label{eq:paa}
    \overline{x}_i = \Big\lceil \frac{w}{n} \Big\rceil \sum_{j=\lfloor\frac{n}{w}(i-1)\rfloor+1}^{\lceil\frac{n}{w}i\rceil}{x_j}\ ,
\end{equation}
where the value $c = \frac{n}{w}$ represents the amplitude of the averaging window and is often referred to as compression rate.

\paragraph{Time series representation: SAX}\label{Symbolic Aggregate approXimation SAX} It is a symbolic representation of time series based on PAA.
The general idea of SAX is to transform the piecewise approximated series $\bar x$ into a ``word'' $\hat x$ composed of symbols from some $\ell$-length alphabet $\AAA=\{\alpha_1, \dots, \alpha_\ell\}$, where each symbol $\alpha_i$ is present with comparable frequency amongst all the words of the dataset.
To do so, the dataset is normalised and PAA-ed. Obtained values are assumed to be distributed according to some chosen distribution, with a Gaussian $\mathcal N(0,1)$ being the typical choice.
A set of $\ell+1$ breakpoints $-\infty = \beta_0 < \beta_1 < \dots < \beta_{\ell-1} < \beta_\ell = \infty$ are then built such that they equally split the chosen distribution in segments of area $\sfrac1\ell$, \ie such that $\int_{\beta_i}^{\beta_{i+1}} \frac1{\sqrt{2\pi}}\exp{(-\sfrac{x^2}2)}dx = \sfrac 1\ell$ when dealing with $\mathcal N(0,1)$.
The PAA approximated values $\bar x_i$ are then converted as
\begin{equation}\label{eq:SAX}
    \hat{x}_i = \alpha_j \hspace{10pt} \text{ if }  \hspace{10pt}\beta_{j-1}\leq \overline{x}_i <\beta_j.
\end{equation}

\paragraph{Time series representation: ESAX}

It represents an extension of the regular SAX where the PAA applied over the normalised dataset retains the minimum and maximum value of each time frame, along with the regular average defined as in~\eqref{eq:paa}.
Resulting values are then converted according to~\eqref{eq:SAX} in a $3\ell$-length word, where each triplet of letters represents minimum, average, and maximum of the time frame.

\paragraph{Similarity measure: Euclidean norm}

\begin{figure*}[t]
\centering
\def\dh{2.7cm}
\includegraphics[height=\dh]{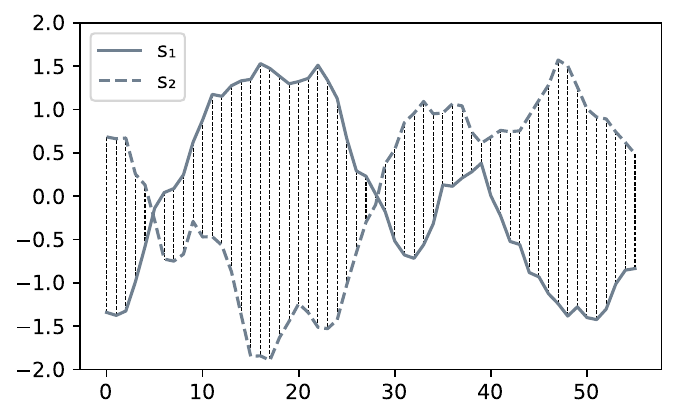}
\includegraphics[height=\dh]{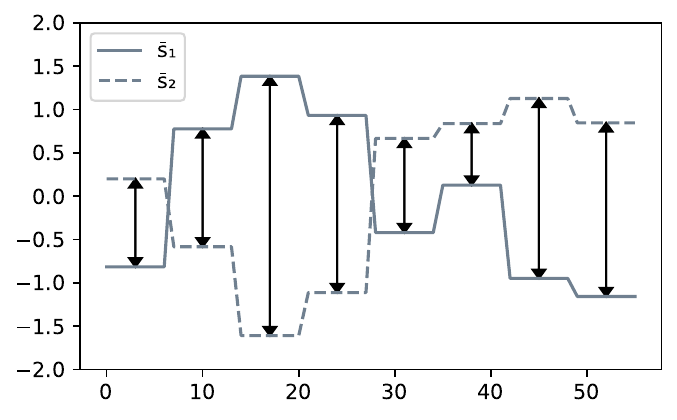}
\includegraphics[height=\dh]{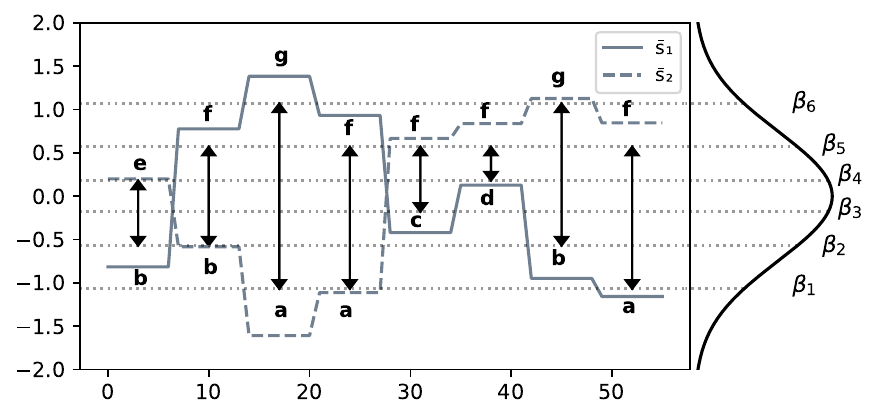}

\caption{
    A visual depiction of the considered distances on 
    (left) $L_2$ over normalized time-series,
    (center) $L_2$ over PAA representation,
    (right) MINDIST over SAX representation,
    when applied on the same couple of series $s_1$ and $s_2$.
}
\label{fig:dist_examples}
\end{figure*}

Given two time series $A = \{a_1,\dots, a_n\}$ and $B = \{b_1, \dots,b_n\}$, the classical $L_2$ norm
\begin{equation}\label{eq:L2}
    L_2(A, B) = \sum_{i=1}^n\sqrt{{a_i}^2-{b_i}^2} 
\end{equation}
can be applied to any numeric time series offering a competitive non-elastic measure of similarity (see also 
Figures~\ref{fig:dist_examples}(left) and~\ref{fig:dist_examples}(center)).
In particular, $L_2$ norm applied on PAA-ed series and on raw series is comparable if rescaled by the size of the time frame.

\paragraph{Similarity measure: MINDIST}

It is a point-wise symmetric symbolic similarity measure defined on symbolic series originally introduced along with SAX representation (Figure~\ref{fig:dist_examples}(right)).
It assesses similarities between two characters $\alpha_i$ and $\alpha_j$ as the smallest distance between the two corresponding intervals, namely
\begin{equation}
    \label{eq:MINDIST}
    \dist(\alpha_i,\alpha_j)=
    \begin{cases}
        \beta_{j-1}-\beta_{i}, & \text{if } i < j-1,\\
        0, & \text{if } |i-j| \leq 1, \\
        \beta_{i-1}-\beta_{j}, & \text{if } i > j+1.
    \end{cases}
\end{equation}

\begin{figure}[t]
\centering
\begin{subfigure}{.48\textwidth}
\centering
  \includegraphics[width=.85\linewidth]{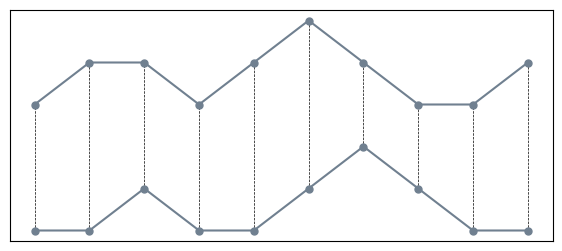}
  \label{fig:dtw1}
\end{subfigure}
\hfill
\begin{subfigure}{.48\textwidth}
\centering
  \includegraphics[width=.85\linewidth]{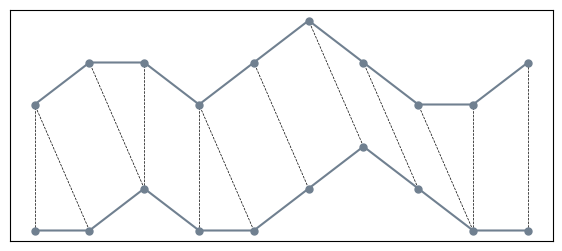}
  \label{fig:dtw2}
\end{subfigure}
\caption{
    Visual difference between Euclidean distance (left) and DTW (right) over a sample time series.
    }
\label{fig:DTW}
\end{figure}

\paragraph{Similarity measure: DTW}
It is a non-linear metric originally proposed in the field of speech recognition and widely adopted in the literature on time-series clustering.
Its widespread use is comparable to that of the $L_2$ norm, but it is usually more accurate, especially when applied to small datasets~\cite{Aghabozorgi2015}.
In fact, in the DTW, each point of the first time series is compared to an arbitrary point of the second one while maintaining the respective ordering, hence resulting in a virtual stretching/compression of the original series (see also Figure~\ref{fig:DTW}).

More formally, let $A =\{a_1,\dots,a_n\}$ and $B=\{b_1,\dots,b_m\}$ be two series (of potentially different length~\cite{ratanamahatana2004everything}), and let $\delta$ be some distance measure defined over the corresponding value domain.
Then DTW (in its iterative definition) evaluates the best alignment, finding a warping path $\pi$~\cite{JMLR:v21:20-091, petitjean2011global} by progressively building the subsequences distance matrix and finding the path of least cost.
Multiple formulations and algorithms have been proposed in the literature over time, offering slight variations and improvements, including normalisation options and rootless versions; in what follows, we adopt the formulation from~\cite{mitsa2010temporal,JMLR:v21:20-091}, \ie

\begin{equation}\label{eq:DTW}
    \DTW(A,B)=\min_\pi\biggl\{\sqrt{\sum_{(i,j)\in\pi}{(a_i-b_j)^2}}\biggl\}\ .
\end{equation}
The choice is motivated by the fact that~\eqref{eq:L2} can be seen as a special case of~\eqref{eq:DTW}, obtained when the best alignment is point-wise.
This makes the two distances intrinsically comparable with one another.
DTW is also a computationally intensive metric to evaluate; hence, in our application, we limit the possible warping path spreading using the Sakoe-Chiba band~\cite{mitsa2010temporal,sakoe1978dynamic}.

\paragraph{Clustering method: $k$-means}
Being one of the most common clustering approaches, $k$-means regroups the objects in clusters around $k$ centroids, which are obtained, in turn, as the mean of the clusters themselves. 
$k$-means is then an iterative deterministic process which converges to a local optimum that depends solely on the initial centroids.
Consequently, different ways of determining the initial conditions were developed, the most adopted being $k$-means\plusplus{}~\cite{arthur2007k}, which progressively assigns initial centroid centres according to probability distributions over the datapoints themselves.

In what follows, centroid evaluation is carried out with unweighted mean on regular and PAA-ed time series, Dynamic Barycenter Averaging (DBA~\cite{petitjean2011global}) on DTW-ed series, and a custom median-based algorithm built on the MINDIST for symbolic series; the latter approach is, to the best of our knowledge, novel and a complete description can be found in Appendix~\ref{app:novel-averaging}.

\paragraph{Clustering method: fuzzy $c$-means}
It can be seen as a variation of regular $k$-means, where cluster membership is somewhat replaced with a notion of affinity, meaning that in fuzzy $c$-means, each data point belongs to a specific cluster with some probability depending on the distance from the cluster centroid.
To alleviate the typical convergence issues which can be found in high dimensional spaces \cite{highdfuzzy}, we adopt the fuzzy $c$-means\plusplus{} \ initialization algorithm \cite{STETCO20157541} with fuzziness coefficient $m=2$.

\paragraph{Clustering method: agglomerative hierarchical clustering analysis (HCA)}
It iteratively builds a hierarchy of the dataset in a bottom-up fashion, starting with a cluster per element and pair-wise merging progressively the clusters containing close elements.
In our experiments, element distance is extended to cluster distance according to the average linkage strategy.
To extract the clusters from the so-formed hierarchy  (\ie cut the dendrogram), we employed a heuristic originally introduced in \cite{Onofri23}, seeking plateaux in the number of $p$-significant clusters to find a stable phase of the clustering process. If no such phase exists, the clustering is deemed a failure.

\paragraph{Clustering evaluation}

To evaluate the results of $k$-means\plusplus{} to the dataset, we chose the number of clusters based on the Silhouette index of the clustering result~\cite{rousseeuw1987silhouettes}, a value in $[-1, 1]$ obtained as the average difference of inter-cluster dissimilarity and intra-cluster dissimilarity.
Conversely, to evaluate the results of the fuzzy $c$-means\plusplus{}, we adopted the PCAES~\cite{wu2005cluster}, consisting of a combination in $[-k, k]$ between a normalized partition coefficient of the fuzzy partition matrix and an exponential separation measure hindering clusters too close to one another.
While a normalization in $[-1, 1]$ seems natural, it is not advised~\cite{wu2005cluster}.
Finally, for what concerns HCA, we selected the valuable outputs under the stability of the $p$-significant clusters.

\medskip
As a final note, it is important to recall that some approaches required hyper-parameters to be fixed a priori (or tuned).
A summary of the five main approaches considered, along with the corresponding hyper-parameter choices, is reported in Table~\ref{tab:dist}.
Here, do mind that DTW is not directly used as PDTW is adopted instead.
This is due to the prohibitive dimensions involved (dataset size and series length), which would yield prohibitive computational costs.

\begin{table*}[]
    \centering
    \caption{Summary of representation and chosen distance measures.}\label{tab:dist}
\begin{tabular}{cccl}
    \toprule
    \textbf{Name} &\textbf{Representation} & \textbf{Similarity} & \textbf{Hyperparameters}\\
    \midrule
     \text{E}&Raw data & Euclidean  &--  \\
     \text{PAAE}&PAA  & Euclidean &$w=144$\\
     \text{SAX}& SAX &MINDIST& $w=144,$ $ a=9$ \\
     \text{ESAX}&ESAX &MINDIST& $w=144,$ $a=9$ \\
     \text{PDTW}&PAA &DTW& $w=144$, \textit{sakoe-chiba} radius $=6$ \\
     \bottomrule
\end{tabular}
\end{table*}

\subsubsection*{Going multivariate}
All the discussed representations, methods, and indices can be naturally extended to multivariate series using different approaches. In particular, our multivariate data is made of a small number of features only (either two or three features, cf.\ Table~\ref{table:multivariate}), so we decided to process the features independently for what concerns representation and evaluation. Similarity scores are hence evaluated on the single sub-series and then summed together.

In the following discussion, we present the PDTW approach only as a proof of concept for the multivariate series, being the most promising amongst the five.

\subsection{Anomaly detection}\label{ssec:Anomaly Detection}

In this section, we detail our approach to anomaly detection by building over the various clustering techniques introduced so far.
In particular, given the daily nature of our data, and provided we would like to create a methodology running on a daily basis to detect and communicate anomalous behaviour, let $\XXX_d$ be the set of time series collected in a given day $d$, purged from trivial anomalies like missing data (cf.\ Section~\ref{ssec:discussing-univariate}).
Then, the idea is to create a suitable set of $\ell$ \textit{anomaly score}s (one per each clustering considered) that can be used to assess how a given time series $s \in \XXX_d$ deviates from the corresponding clusterings and process such scores to build a set $\AAA_d \subset \XXX_d$ of the ``potential anomalous'' data.

In details, for each considered hard clustering ($k$-means, HCA), the distance from the closest centroid typically provides a good proxy of the abnormality of the datum, hence forming a simple yet effective anomaly score.
Conversely, when considering soft clustering (fuzzy $c$-means), we use the classical approach consisting of averaging the distance from each centroid weighted on the probability of belonging to the corresponding cluster.

The resulting set of $\ell$ anomaly scores can then be collected in a $\ell$-rows $|\XXX_d|$-columns matrix $A = (a_{i, j})$ and processed via the following two aggregated scores:
\begin{description}
    \item[Aggregated Average (AGG)] which row-rescales the matrix $A$ in the interval $[0, 1]$ (\ie apply the (0, 1)-normalisation on each score) and averages the resulting scores (\ie perform the average-by-column).
    In other words, the aggregated score for the $j$-th series is obtained as
    \begin{equation}
        \text{AGG}(s_j) = \frac1\ell\sum_{i=1}^\ell \frac{a_{i, j}}{\max\{a_{i, \cdot}\} }
    \end{equation}
    forming an `average distance' in $[0, 1]$ from the `normality' (the higher, the more anomalous).
    
    \item[Positional Ranking (POS)] which applies the rank-transformation by row on $A$ (\ie maps score in $[1, \dots, |\XXX_d|]$ from the higher to the lower value) before evaluating the AGG as before.
    In other words, POS rescale scores in $\{0, \sfrac{1}{|\XXX_d|-1}, \dots,$ $\sfrac{|\XXX_d|-2}{|\XXX_d|-1}, 1\}$ before evaluating an average score, hence providing an anomalous ranking in $[0, 1]$ (the lower, the more anomalous).
\end{description}

Classically, these scores can be then used to discriminate anomalies depending on some fixed thresholds.
However, it is important to note that the normalisation is performed day-wise, \ie for each daily dataset $\XXX_d$ , different values $\max\{a_{i, \cdot}\}$ and $\min\{a_{i, \cdot}\}$ are chosen; this is due to the online nature of the process, where data from the future days are not available, making it impossible to define a fixed minimum or maximum.
As a consequence, series from different days are not directly comparable, making it difficult to define a single threshold value. 

Hence, to finally process a single series $s$, we follow the same rationale of POS, adopting a ranking approach which is, in principle, agnostic of anomalies.
Let $2k$ be the number of potential anomalies to report daily (in our experiments $k = 3$, $2k \sim \sfrac{5}{100}|\XXX_d|$).
Then, two initial sets of daily anomalies are formed as the most relevant $k$ anomalies according to AGG and POS (\ie, the $k$ top scoring for AGG and $k$ bottom scoring for POS), which we will call \textsc{topAGG} and \textsc{topPOS}, respectively.

Exploiting the intrinsic temporal feature of the datasets $\XXX_d$, we can further analyse $\YYY_d$ in terms of the anomalies detected in the previous days $\YYY_{d-1}, \dots \YYY_{d-h}$, for some $h\in \mathbb N$.
In fact, it is reasonable that sensor anomalies might repeat within different days, making them more worthy of attention.
To combine this piece of information with the intrinsic rank provided by AGG and POS, we define the final confidence score $\mathfrak c(s)$ for a series $s \in \XXX_d$ in $[0, 3k]$ as the sum of the two ranks in \textsc{topAGG} and \textsc{topPOS} (if present, in inverted order from $k$ to $1$) plus $k$ if $s \in \YYY_{d-t}, t \in [1, h]$, for at least $g \leq h$ days.

The confidence score $\mathfrak c$ can then be either used \textit{as-is} or it can be classified in grades of confidence/severity.
In our experiments, \eg, we classified anomalies as \textit{mild} when $\mathfrak c \in [1,k]$, \textit{moderate} when $\mathfrak c \in (k,2k]$, and \textit{severe} when $\mathfrak c \in (2k,3k]$.
As a reference, we also set $h=7$ and $g=2$, meaning that \textit{severe} anomalies can occur only when an anomaly is detected at least three times within the last week.

\section{Results}\label{sec:results}
In this section, we firstly present some results about the clustering of the traffic data introduced in Section~\ref{sec:dataset}, followed by an analysis of traffic anomalies and sensor failure detection.
Being the dataset unlabelled, no direct ground truth is available to validate the presented results.
However, the consistency of clustering indicators provides a reliable proxy for assessing the effectiveness of different methodologies.
Furthermore, the results were validated through expert knowledge from stakeholders at Autostrade Alto Adriatico S.p.A.
Finally, during a six-month trial phase, the anomaly detection technique successfully identified real anomalies with a low false positive rate ($<10\%$), with no human intervention required.

\subsection{Clustering results}\label{ssec:clustering-results}
We present the clustering results divided according to the approach for the sake of readability.
 
\begin{figure*}[t]
    \centering
    \hfill
    \includegraphics[width=0.49\linewidth]{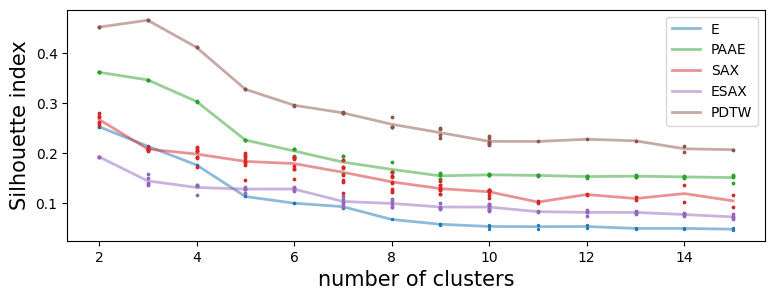}
    \hfill
    \includegraphics[width=0.49\linewidth]{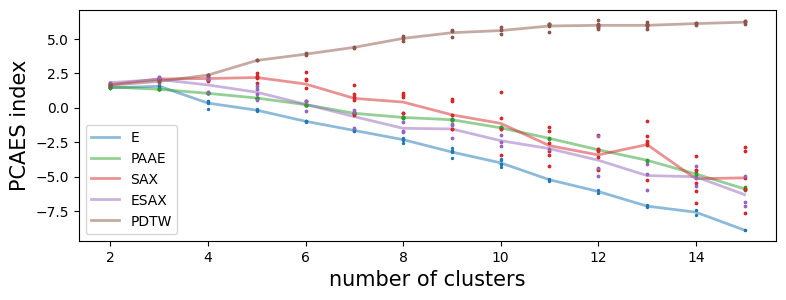}
    \hfill
    \caption{
        Plot of the Silhouette index for $k$-means (left) and PCAES index for fuzzy $c$-means (right) applied to the 5 representation--similarity couples (cf.\ Table~\ref{tab:dist}).
    }
    \label{fig:Sil_PCAES_pres}
\end{figure*}

\paragraph{$k$-means}
Having no information \textit{a priori} on the correct number $k$ of clusters, we tried to identify a suitable value by making use of the Silhouette index.
As depicted in Figure~\ref{fig:Sil_PCAES_pres} (left), the index shows a descending trend whichever representation is chosen (but the PDTW), prompting the less the cluster number the better performance.
While the Silhouette score is not a perfect indicator, this suggests that $k$-means is not able to separate the dataset effectively.

In general, by observing the results, elements in the dataset seem to lack a clear separation in different groups, with data evenly distributed in the space (although with different densities).
This, combined with the dataset dimension, motivates the reduced effectiveness of the $k$-means approach, which aims for a clear-cut separation of the data.

\paragraph{Fuzzy $c$-means} 

\begin{figure*}[t]
\centering
\hfill
\begin{subfigure}{.19\textwidth}
  \centering
  \includegraphics[width=.99\linewidth]{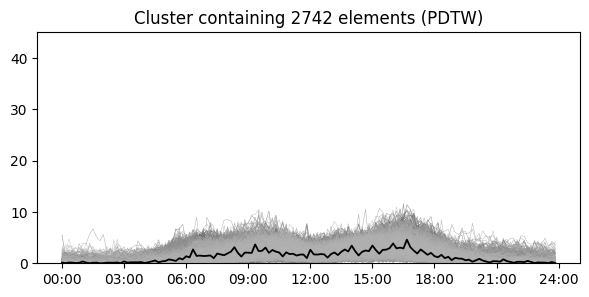}%1
\end{subfigure}
\hfill
\begin{subfigure}{.19\textwidth}
  \centering
  \includegraphics[width=.99\linewidth]{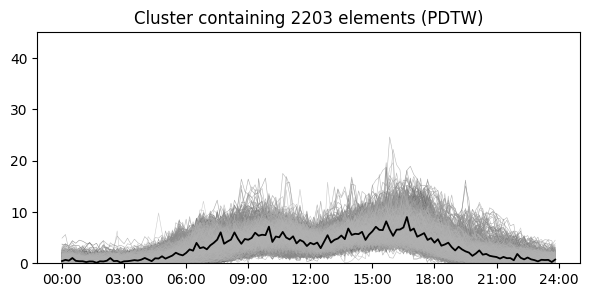}%2
\end{subfigure}
\hfill
\begin{subfigure}{.19\textwidth}
  \centering
  \includegraphics[width=.99\linewidth]{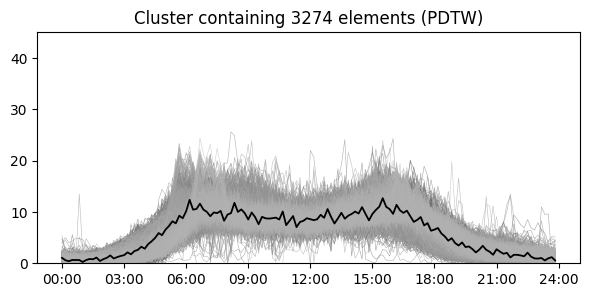}%3
\end{subfigure}
\hfill
\begin{subfigure}{.19\textwidth}
  \centering
  \includegraphics[width=.99\linewidth]{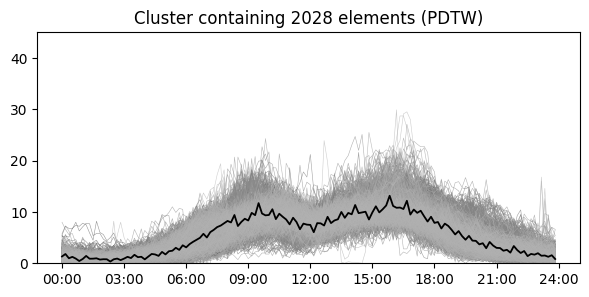}%4
\end{subfigure}
\hfill
\begin{subfigure}{.19\textwidth}
  \centering
  \includegraphics[width=.99\linewidth]{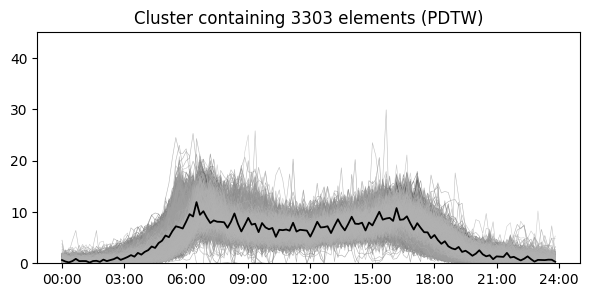}%5
\end{subfigure}
\hfill

\hfill
\begin{subfigure}{.19\textwidth}
  \centering
  \includegraphics[width=.99\linewidth]{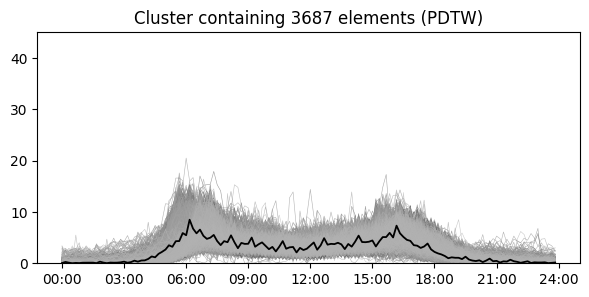}%6
\end{subfigure}
\hfill
\begin{subfigure}{.19\textwidth}
  \centering
  \includegraphics[width=.99\linewidth]{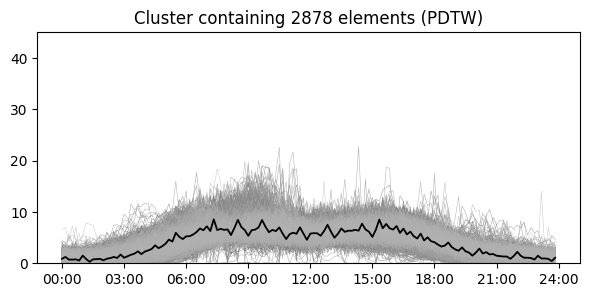}%7
\end{subfigure}
\hfill
\begin{subfigure}{.19\textwidth}
  \centering
  \includegraphics[width=.99\linewidth]{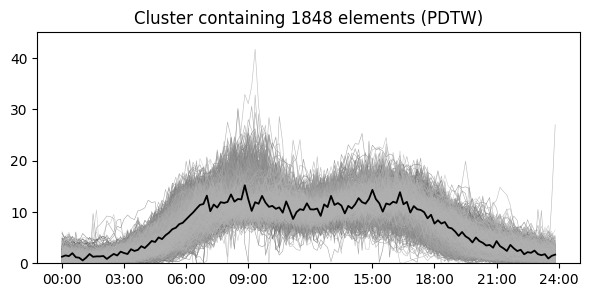}%8
\end{subfigure}
\hfill
\begin{subfigure}{.19\textwidth}
  \centering
  \includegraphics[width=.99\linewidth]{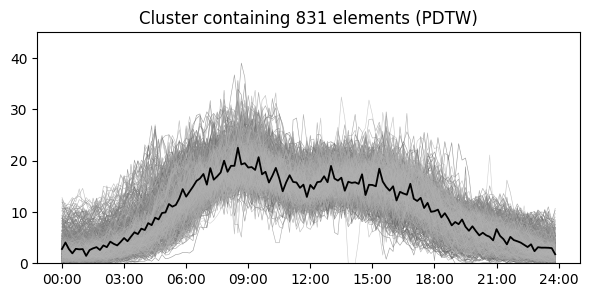}%9
\end{subfigure}
\hfill
\begin{subfigure}{.19\textwidth}
  \centering
  \includegraphics[width=.99\linewidth]{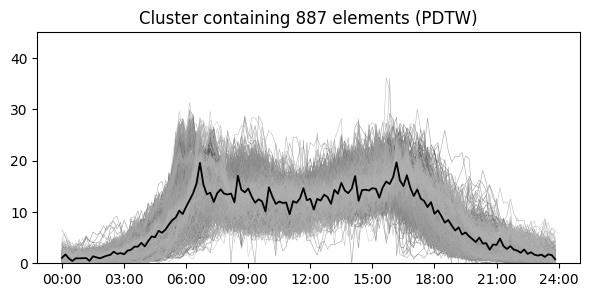}%10
\end{subfigure}
\hfill

\hfill
\begin{subfigure}{.19\textwidth}
  \centering
  \includegraphics[width=.99\linewidth]{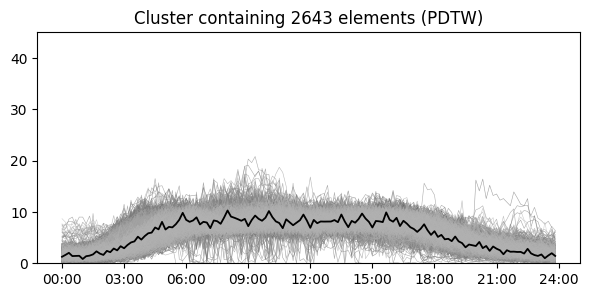}%11
\end{subfigure}
\hfill
\begin{subfigure}{.19\textwidth}
  \centering
  \includegraphics[width=.99\linewidth]{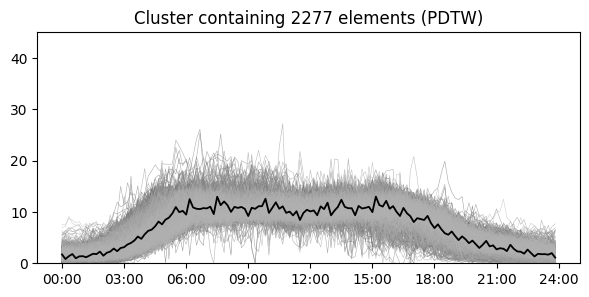}%12
\end{subfigure}
\hfill
\begin{subfigure}{.19\textwidth}
  \centering
  \includegraphics[width=.99\linewidth]{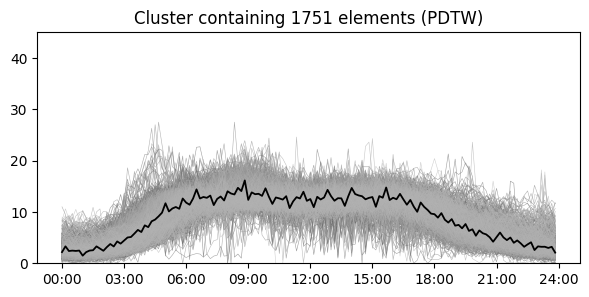}%13
\end{subfigure}
\hfill
\begin{subfigure}{.19\textwidth}
  \centering
  \includegraphics[width=.99\linewidth]{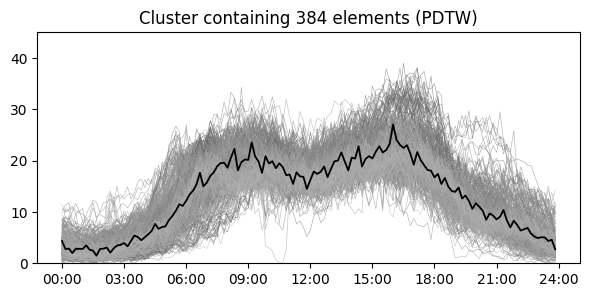}%14
\end{subfigure}
\hfill
\begin{subfigure}{.19\textwidth}
  \centering
  \includegraphics[width=.99\linewidth]{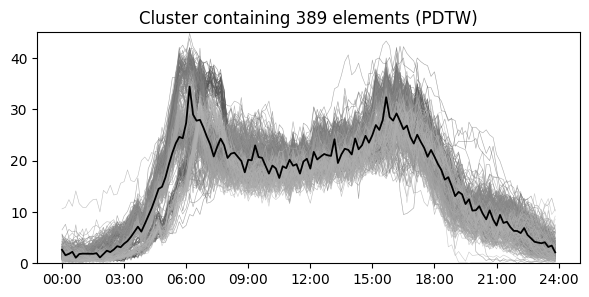}%15
\end{subfigure}
\hfill
\caption{
    Fuzzy $c$-means results of the PDTW representation when $c=15$.
    Each panel represents a series (in grey) assigned to a cluster, with the corresponding centroid overlayed in black.
}
\label{fig:fuzzy15dtw}
\end{figure*}

Analogous to $k$-means, we adopted $\PCAES$ to identify a suitable number $c$ of clusters.
As reported in Figure~\ref{fig:Sil_PCAES_pres} (right), most of the approaches present low PCAES values, which decrease to negative values when $c$ increases, deeming their ineffectiveness.
Conversely, PDTW representation offers promising results with monotone increasing values of PCAES.
In particular, $c=15$ represents a sweet-point after which the index reaches a plateau (values not shown for conciseness).
Figure~\ref{fig:fuzzy15dtw} shows the 15 resulting clusters along with the corresponding centroids.
We can see how the division is mostly driven by the series ``height'', but the high number of clusters helps distinguish different ``shapes'' sharing similar averages.

\medskip

It is worth noting that in both $k$-means and fuzzy $c$-means the series ``height'' (average flux) seems to affect the clustering results predominantly, in particular if compared to the actual series ``shape''.
This is particularly evident in the best clustering of $k$-means (PDTW, $k=3$, cf.\ Figure~\ref{fig:Sil_PCAES_pres} left, clusters not shown for conciseness), where three clusters are formed collecting three orders of magnitude of the flux.
While this does not represent a wrong result \textit{per se}, it highlights how the distance selection can affect the direction the analysis takes.

Similar considerations can be drawn for the pairing of representation space and similarity score. In fact, the suboptimal results achieved by SAX and ESAX are not surprising in the context of partitioning methods:
symbolic representation discretizes the space, hence hampering the ability of $k$-means and fuzzy $c$-means to slowly move the cluster centres in the space until a suitable position is found, to the point that some initial parameter configurations failed to converge.

\paragraph{Hierarchical methods}
As per the previous methods, we applied the hierarchical approach over all the introduced representations.
However, non-normalized representation proved to be ineffective due to the inhomogeneity in the average flux, with clusters agglomerating low-flux series in a few clusters while ignoring most of the higher extremes.

\begin{figure*}
    \centering
    \includegraphics[width=.38\linewidth]{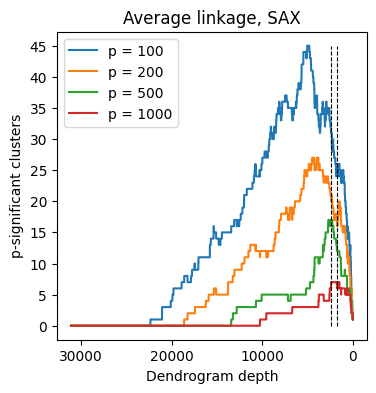}
    \includegraphics[width=.38\linewidth]{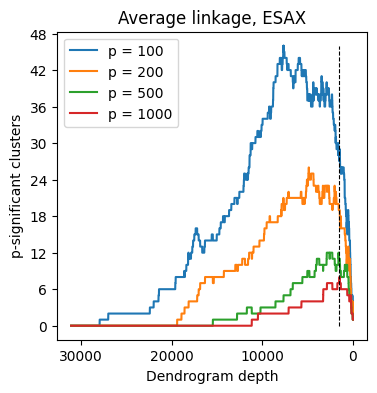}
    \hfill
    \caption{
        Plot of the number of $p$-significant clusters (clusters with more than $p$ series assigned) obtained during the agglomeration procedure applied on SAX (left) and ESAX (right) representations. 
        Dendrogram cuts detected by the heuristic are highlighted with a black line.
    }
    \label{fig:hsaxhesax}
\end{figure*}

\begin{figure*}[t]
\centering
\begin{subfigure}{.24\textwidth}
  \centering
  \includegraphics[width=.99\linewidth]{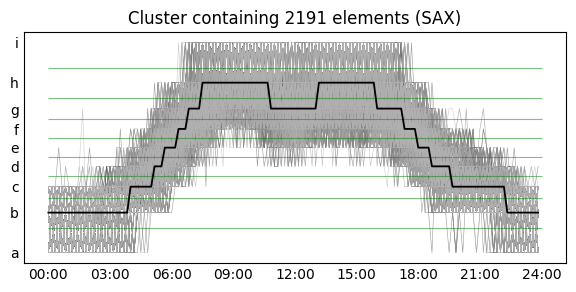}
\end{subfigure}
\hfill
\begin{subfigure}{.24\textwidth}
  \centering
  \includegraphics[width=.99\linewidth]{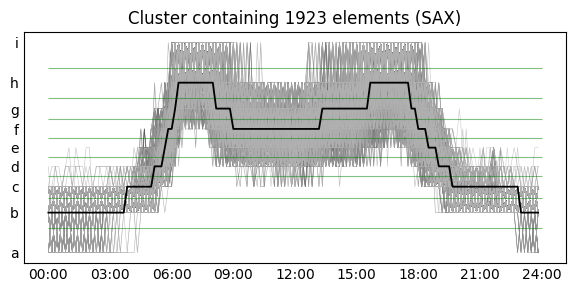}
\end{subfigure}
\hfill
\begin{subfigure}{.24\textwidth}
  \centering
  \includegraphics[width=.99\linewidth]{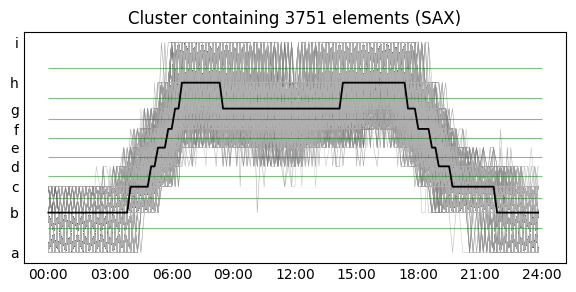}
\end{subfigure}
\hfill
\begin{subfigure}{.24\textwidth}
  \centering
  \includegraphics[width=.99\linewidth]{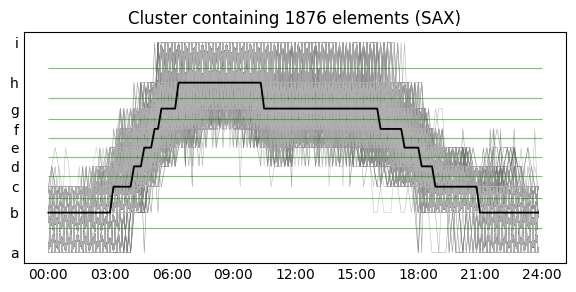}
\end{subfigure}
\hfill

\begin{subfigure}{.24\textwidth}
  \centering
  \includegraphics[width=.99\linewidth]{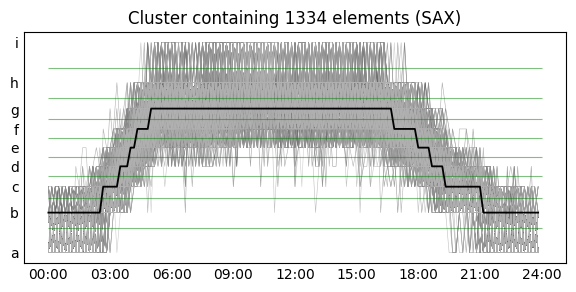}
\end{subfigure}
\begin{subfigure}{.24\textwidth}
  \centering
  \includegraphics[width=.99\linewidth]{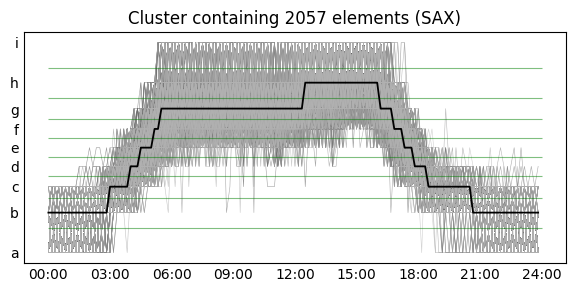}
\end{subfigure}
\begin{subfigure}{.24\textwidth}
  \centering
  \includegraphics[width=.99\linewidth]{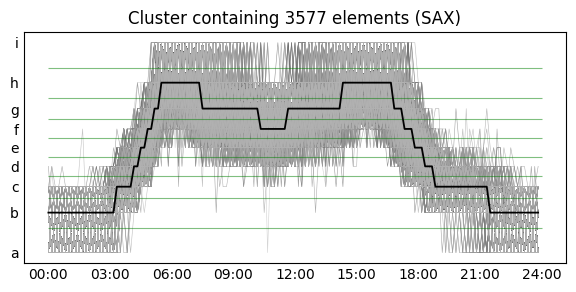}
\end{subfigure}
\hfill

\caption{
    Clustering results obtained cutting the dendrogram of SAX representation (cf.\ Figure~\ref{fig:hsaxhesax} left, right cut).
    Each panel represents series (in grey) assigned to a large cluster ($>1000$ series), with the corresponding centroid overlayed in black (cf.\ Algorithm~\ref{alg:SAX_mean}).
}
\label{fig:hier7sax}
\end{figure*}

For the remaining representations, we adopted the heuristic described in Section~\ref{ssec:clustering-method} to locate suitable dendrogram cuts (cf.\ Figure~\ref{fig:hsaxhesax}).
We also decided to filter clusters smaller than $1000$ elements in order to exclude outliers from the clustering.
Figure~\ref{fig:hier7sax} presents the results obtained with SAX representation in terms of seven clusters.
Differently from previous approaches, due to the applied normalisation and the outlier removal, we can clearly distinguish different shapes emerging while flux magnitude is almost completely invisible.

ESAX results are similar in shape and conclusions and are omitted for the sake of conciseness.

\paragraph{Multivariate analysis results}
Considering classes and lanes individually allowed obtaining numerous different insights and results, unveiling patterns previously hidden by the natural aggregation considered.
In what follows, we report some of the most interesting conclusions for each approach with no claim of completeness.

\begin{figure*}[t]
\centering
\begin{subfigure}{.49\textwidth}
  \centering
  \includegraphics[width=.99\linewidth]{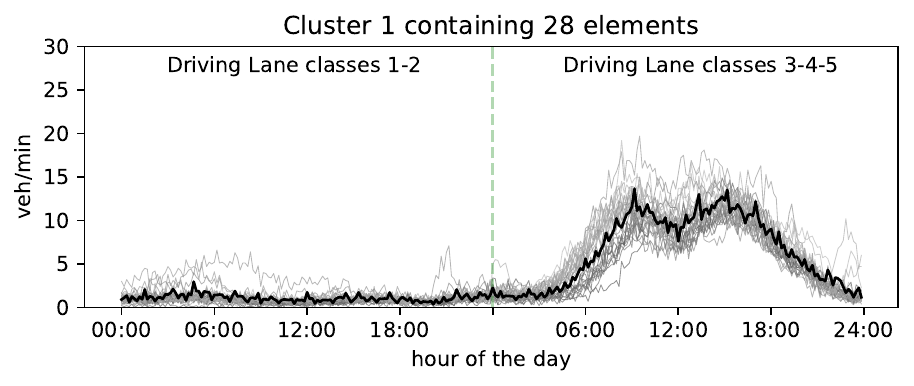}
\end{subfigure}
\hfill
\begin{subfigure}{.49\textwidth}
  \centering
  \includegraphics[width=.99\linewidth]{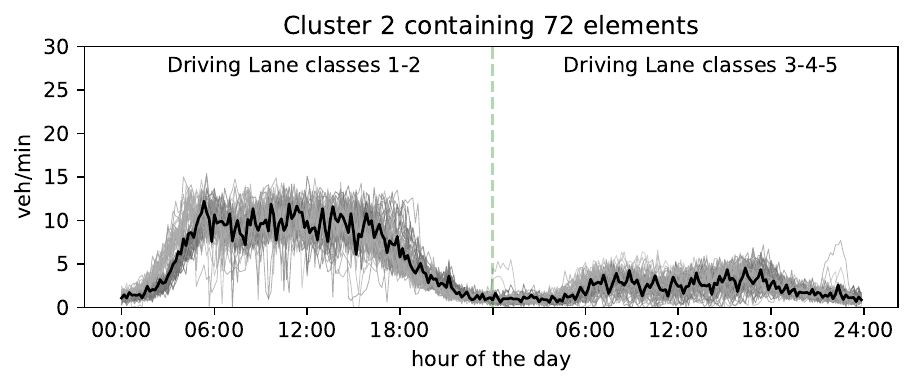}
\end{subfigure}
\hfill

\hfill
\begin{subfigure}{.49\textwidth}
  \centering
  \includegraphics[width=.99\linewidth]{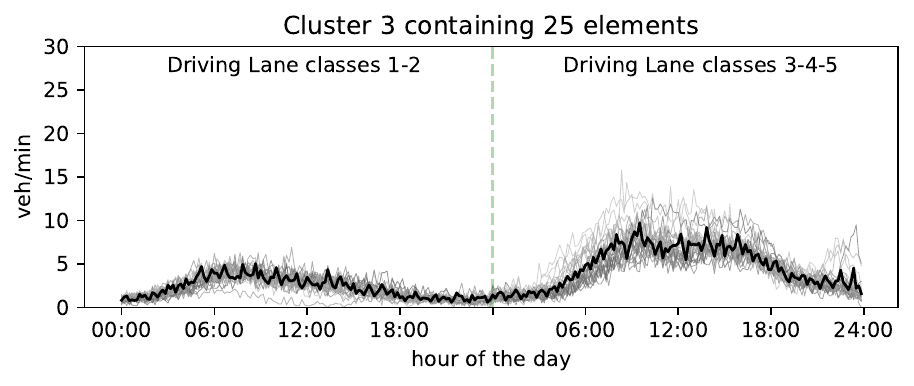}
\end{subfigure}
\hfill
\begin{subfigure}{.49\textwidth}
  \centering
  \includegraphics[width=.99\linewidth]{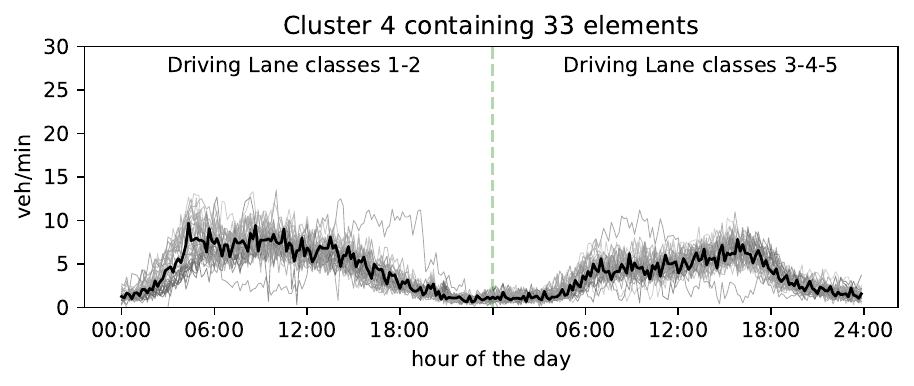}
\end{subfigure}
\hfill
\caption{
    Multivariate (T1) fuzzy $c$-means results of the PDTW representation when $c=4$.
    Each panel represents series (in grey) assigned to a cluster, with the corresponding centroid displayed in black.
    Based on the calendar days: Cluster 1 (top left) collects 1 (of 24) Saturday, all Sundays (23/23) and national holidays (4/4) contained in the dataset; Cluster 3 (bottom left) collects remaining Saturdays (23/24).
}
\label{fig:multi_result}
\end{figure*}

As an example, every clustering obtained with at least three clusters over type T1 (cf.\ Table~\ref{table:multivariate}) was able to almost perfectly sort out weekdays from weekends.
When increasing the number of clusters, some combinations also divided Saturdays from Sundays, with holidays typically grouped with Sundays.
This behaviour is intrinsic with the difference in traffic composition, with \eg heavy vehicles nearly missing on non-working days.
Figure~\ref{fig:multi_result} presents an example of clustering results achieved on T1 with 4 clusters showcasing this feature.

Approach T2 and T3 both obtained similar results but with lower accuracy in distinguishing between weekdays and holidays; in particular, the aggregation of \textit{light vehicles} on both lanes tended to create outlier clusters presenting extremely high light flux (mostly holidays, but without separation between Saturdays and Sundays).

\subsection{Anomaly detection results}\label{ssec:anomaly-results}
To build our anomaly detector, we selected from the previous section the three most promising clustering techniques, namely (i) the fuzzy clustering combined with PDTW (cf.\ Figure~\ref{fig:fuzzy15dtw}), (ii) the hierarchical clustering with SAX (cf.\ Figure~\ref{fig:hier7sax}), and (iii) the hierarchical clustering with ESAX.
This provided us with $\ell=3$ anomaly scores for a given series, each one providing a relevant contribution being tuned on different characteristics in the series.
Particularly, partitioning on PDTW is more focused on the flux values (series height), HCA on SAX on the shape of the series, whereas HCA on ESAX on single-point anomalies (which go unnoticed by the not-Extended counterpart).

\begin{figure*}[]
\centering

\begin{subfigure}{.32\textwidth}
  \centering
  \includegraphics[width=\linewidth]{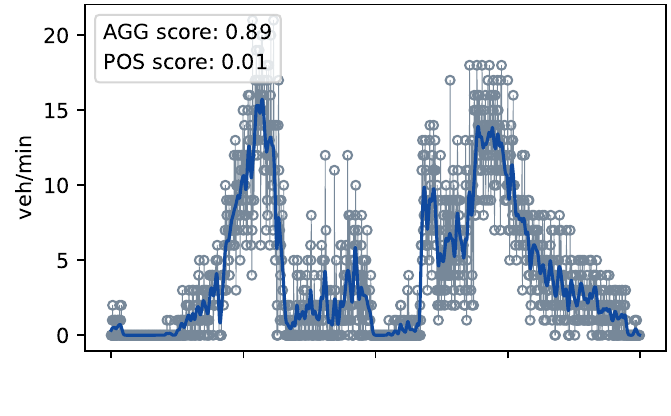}
\end{subfigure}
\hfill
\begin{subfigure}{.32\textwidth}
  \centering
  \includegraphics[width=\linewidth]{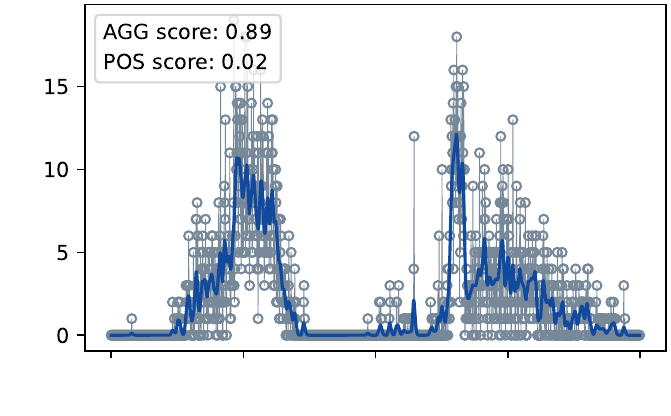}
\end{subfigure}
\hfill
\begin{subfigure}{.32\textwidth}
  \centering
  \includegraphics[width=\linewidth]{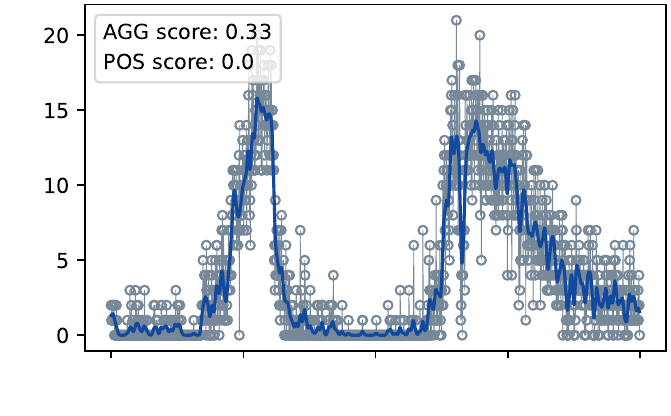}
\end{subfigure}

\begin{subfigure}{.32\textwidth}
  \centering
  \includegraphics[width=\linewidth]{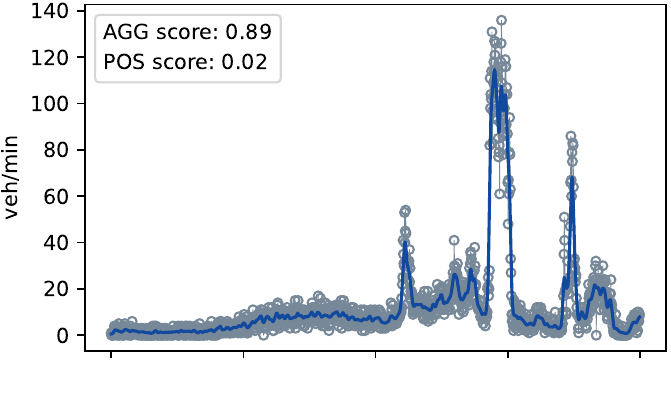}
\end{subfigure}
\hfill
\begin{subfigure}{.32\textwidth}
  \centering
  \includegraphics[width=\linewidth]{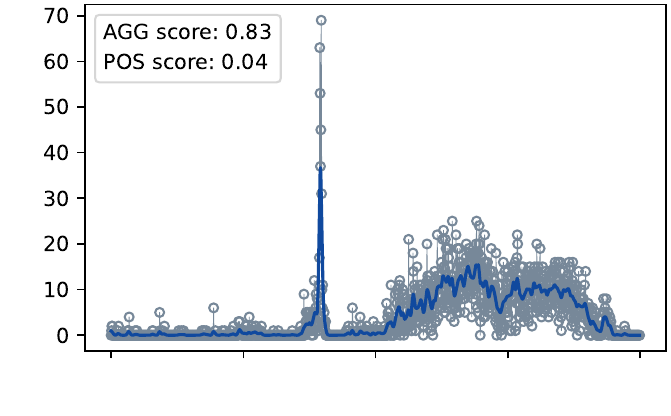}
\end{subfigure}
\hfill
\begin{subfigure}{.32\textwidth}
  \centering
  \includegraphics[width=\linewidth]{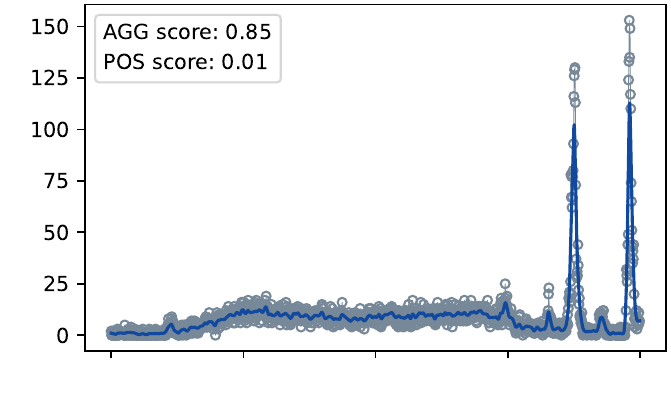}
\end{subfigure}

\begin{subfigure}{.32\textwidth}
  \centering
  \includegraphics[width=\linewidth]{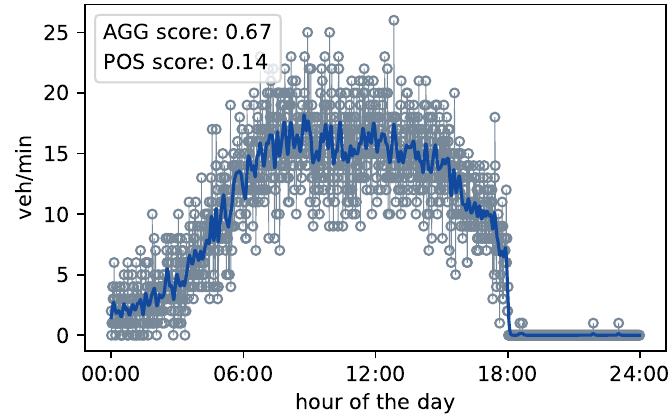}
\end{subfigure}
\hfill
\begin{subfigure}{.32\textwidth}
  \centering
  \includegraphics[width=\linewidth]{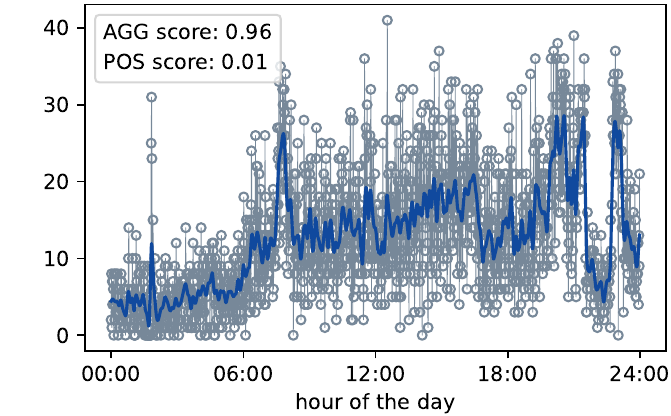}
\end{subfigure}
\hfill
\begin{subfigure}{.32\textwidth}
  \centering
  \includegraphics[width=\linewidth]{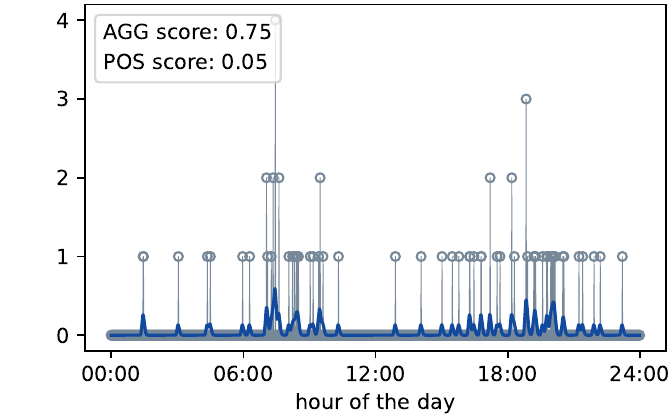}
\end{subfigure}

\caption{
    Example of daily anomalies found using the proposed anomaly detection techniques.
    The AGG and POS scores are reported in the images.
    Do notice that, for the sake of readability, $y$-axes are not homogeneous.
    Also, scores are day-dependent, so they are not directly comparable.
    The top row reports three probable traffic anomalies caused by long congestion events.
    The central row depicts three anomalies caused by sensor failures communicating unfeasibly high flux spikes.
    The bottom row presents three different kinds of events: (left) a sensor abruptly stopping reporting vehicles, (centre) unstable traffic conditions with abnormal flux (in particular at night time), and (right) a very low traffic reading either ascribed to sensor malfunctioning or to lane closure.
} 
\label{fig:anomalies}
\end{figure*}

To combine the different anomaly scores, we implemented both the AGG and POS approaches introduced in Section~\ref{ssec:Anomaly Detection}. The results were empirically proven to be good at identifying the most anomalous series during a single day. 
Figure~\ref{fig:anomalies} provides a few representative sample results.

\medskip

We also conducted an initial anomaly detection analysis using the results from multivariate clustering.
To understand its potentialities, we ranked the multivariate datasets based on the reconstruction error.
Since the original datasets were pruned of trivial anomalies, the focus shifted to the additional insights offered by this method on normal-looking series.

\begin{figure*}
    \centering
    \includegraphics[width=1\linewidth]{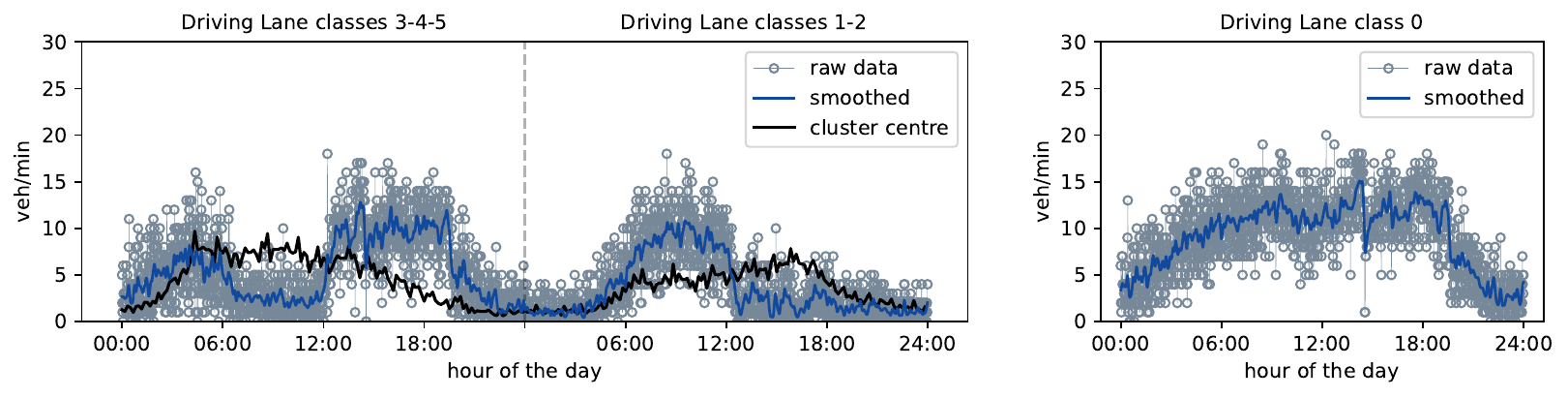}
    \caption{
        Example of a traffic anomaly detectable with multivariate analysis only.
        Flux of light and heavy vehicles (left panel) surprisingly sum to a regular total flux (right panel).
        Anomaly is detected with fuzzy $c$-means over PDTW representation (cf.\ Figure~\ref{fig:multi_result}, Cluster 4).
    }
    \label{fig:multivariate-anomaly}
\end{figure*}

As expected, the multivariate approach highlighted different characteristics compared to the univariate approach, emphasizing differences between classes.
For instance, Figure~\ref{fig:multivariate-anomaly} shows a series that consistently ranks among the furthest from the corresponding centroid in the multivariate $k$-means analysis despite being pretty ``normal'' if considered in the classical univariate analysis.
Here, in fact, the light vehicles are almost absent in the afternoon, while the heavy counterpart is almost absent in the morning, hence composing fluxes that both present a particular shape which, surprisingly, balances out in the aggregated flux.
A deeper analysis reveals the day being a post-holiday.

While such cases do exist, their presence in the dataset seems to be limited, with no particular applicative appeal.
As a consequence, we decided not to proceed with further analysis.

\section{Conclusions and future work}\label{sec:conclusions}

In this paper, we introduced a comprehensive framework for analysing time-series data using clustering techniques, detailing the essential preprocessing steps, similarity measures, clustering methodologies, and evaluation techniques. We draw our focus on the interaction between these components and the necessary adaptations required to obtain meaningful results, such as the symbolic averaging method described in Appendix~\ref{app:novel-averaging}.

We applied our methodology to a real-world traffic dataset, demonstrating the strengths and limitations of various clustering techniques and how dataset characteristics influence outcomes. Additionally, we explored anomaly detection, both as a theoretical framework and through a practical implementation, leveraging clustering results for an efficient identification of irregular traffic patterns.

Our results have shown that, for our dataset, applying fuzzy $c$-means clustering using PDTW is the best choice in terms of general separation of the dataset, when the absolute values of the series are taken into consideration. 
Conversely, by introducing a normalization step, combining SAX/ESAX representations with HCA yields the best results when judged by the ability to discern shapes. 
While these final considerations are valid specifically for our dataset, we also gained more general conclusions, such as the incompatibility between symbolic representations and partitioning clustering.

Our findings offer valuable insights into creating a pipeline for time-series clustering and anomaly detection, but they represent only an initial step. Future research should expand on the multivariate approach and refine anomaly detection, focusing on differentiating between sensor malfunctions and genuine traffic anomalies. Additionally, further investigation is required to establish which clustering methods yield the most reliable results across different traffic scenarios.

\addtocontents{toc}{\SkipTocEntry}
\subsection*{Author contribution}

D.M.\ prepared the first draft, performed the literature review, developed the code, conducted initial testing, extracted the results, and created the graphics.
E.O.\ initially analysed the dataset for clustering and anomaly detection, supervised the developments, provided the initial code, and contributed to the literature review, writing, and result analysis and discussion.
E.C.\ supervised the overall research, conceptualised the study, secured funding, and managed the collaboration with the data-providing company.
All authors validated the results with the assistance of the data-providing company.
All authors reviewed and approved the final manuscript.

\addtocontents{toc}{\SkipTocEntry}
\subsection*{Acknowledgements}

The authors want to warmly thank Andrea Lisi for his support in analysing and understanding data, as well as improving the quality of the labelled dataset.
The authors also want to thank Paolo Ranut, who supported this research on behalf of the motorway company Autostrade Alto Adriatico S.p.A.

All authors are members of the Gruppo Nazionale Calcolo Scientifico - Istituto Nazionale di Alta Matematica (GNCS-INdAM).

\addtocontents{toc}{\SkipTocEntry}
\subsection*{Fundings}

This work was partially funded by Autostrade Alto Adriatico S.p.A.

E.C.\ would like to thank the Italian Ministry of University and Research (MUR) for supporting this research with funds coming from PRIN Project 2022 PNRR (No. 2022XJ9SX, entitled ``Heterogeneity on the road - Modeling, analysis, control'').

This study was carried out within the Spoke 7 of the MOST -- Sustainable Mobility National Research Center and received funding from the European Union Next-Generation EU (PIANO NAZIONALE DI RIPRESA E RESILIENZA (PNRR) – MISSIONE 4 COMPONENTE 2, INVESTIMENTO 1.4 – D.D. 1033 17/06/2022, CN00000023). This manuscript reflects only the authors' views and opinions. Neither the European Union nor the European Commission can be considered responsible for them.

\bibliographystyle{abbrv}
\bibliography{biblio}

\appendix

\section{Averaging symbolic series}\label{app:novel-averaging}
In this appendix, we introduce an approach which is, to the best of our knowledge, previously unpublished to evaluate the centroid of a given cluster of symbolic representations $C=\{\hat{s}_1,\dots,\hat{s}_m\}$ (with each series $\hat{s}_j = (x_1^{(j)},\dots,x_w^{(j)})$) relatively to MINDIST~\eqref{eq:MINDIST}. Algorithm~\ref{alg:SAX_mean} summarise the process.

\begin{algorithm}
\caption{\texttt{Symbolic\_Average(c)}}\label{alg:SAX_mean}
\KwData{A set $\{\hat{s}_1,\dots,\hat{s}_m\}$ of symbolic $w$-length series $\hat{s}_i$ of alphabet $\AAA$}
\KwResult{An average symbolic series $\mu$.}
\For{$i = 1,2,3,\dots,w$}{
    \For{$\alpha$ in $\AAA$}{
         $n_\alpha =  \boldsymbol\#\{x_i^{(j)}$ \textbf{for} $j \in \{1,\dots,m\}$ \textbf{if} $x_i^{(j)}=\alpha\} $\;
    }
    $\mu_i = \argmin_{\beta \in \AAA}\Big({\sum_{\alpha \in \AAA} \big(n_\alpha\cdot(\dist(\alpha,\beta))^2}\big)\Big)$\;
    }
\end{algorithm}

The fundamental part of the proposed algorithm lies in the analogy with the classical version in $\mathbb R^w$.
In fact, the approximation of the centroid $\mu = \{\mu_1,\dots,\mu_w\}$, which is typically obtained by averaging the time series, can be reformulated as the search problem of finding $\mu \in \mathbb{R}^n$ as
\begin{equation}
\centering
\mu = \argmin_{r}\frac{1}{m}\sum_{i=1}^{m}{\Ldue(s_i,r)^2}
\end{equation}
Given a set of symbolic representations $\{\hat{s}_1,\dots,\hat{s}_m\}$ and the distance measure MINDIST, finding the sequence $r=\{r_i,\dots,r_w\}$ such as
    \begin{equation}\label{obj_avg_sax}
    r = \argmin_{s\in \mathcal{S}}\frac{1}{m}\sum_{i=1}^{m}{\big(\mindist(\hat{s}_i,s)\big)^2}
    \end{equation}
where $\mathcal{S}$ is the set of all possible symbolic sequences $s=\{x_1,\dots,x_w\}$ of length $w$, can be seen by MINDIST definition as
\begin{equation} \label{eq:obj_avg_sax_2}
    r = \argmin_{s\in \mathcal{S}}\sum_{i=1}^{m}{\biggl(\sqrt{\frac{n}{w}\sum_{j=1}^{w}\big(\dist(\hat{x}_j^{(i)},{x}_j\big)^2}\biggl)^2}
\end{equation}
which equals
\begin{equation} \label{eq:final_sax_avg}
    r = \argmin_{s\in \mathcal{S}}\sum_{i=1}^{m}{{\Big(\frac{n}{w}}{\sum_{j=1}^{w}\big(\dist(\hat{x}_j^{(i)},x_j)\big)^2}\Big)}
    = \argmin_{s\in \mathcal{S}}\sum_{i=1}^{m}{\Big({\sum_{j=1}^{w}\big(\dist(\hat{x}_j^{(i)},x_j)\big)^2}\Big)}\ .
\end{equation}
Since every $\dist()$ computation is independent, we can rewrite~\eqref{eq:final_sax_avg} as
\begin{equation}
        r
    =\argmin_{s\in \mathcal{S}}\sum_{j=1}^{w}{\big({\sum_{i=1}^{m}\big(\dist(\hat{x}_j^{(i)},x_j)\big)^2}\big)}.
\end{equation}

Now, every $j$ computation is independent of the other, which means that every component of the first sum can be optimized separately. Hence, the $j$-th component of $r$ equals the letter $\beta \in \AAA$ that minimizes the internal sum, \ie
\begin{equation}\label{eq:final_final_sax_avg}
    r_j=\argmin_{\beta\in \AAA}{\sum_{i=1}^{m}\big(\dist(\hat{x}_j^{(i)},\beta)\big)^2}.
\end{equation}
Since both $\hat{x}_j^{(i)}$ and $\beta$ are in the same finite alphabet $\AAA$, the number of possible distances is finite; hence, we can rewrite~\eqref{eq:final_final_sax_avg} as
\begin{equation}
\label{eq:final_final_final_sax_avg}
    r_j=\argmin_{\beta\in \AAA}{\sum_{\alpha \in \AAA}\Big(n_\alpha\cdot\big(\dist(\alpha,\beta)\big)^2\Big)}, 
\end{equation}
where $n_\alpha$ is the cardinality of $\{\hat{x}_j^{(i)}\text{ } ,i=1,\dots,m$ $|$ $\hat{x}_j^{(i)}=\alpha\}$, matching the thought process behind Algorithm~\ref{alg:SAX_mean}. Using~\eqref{eq:final_final_final_sax_avg} instead of~\eqref{eq:final_final_sax_avg} reduces the number of distance computations, although increasing the total number of simple operations.

\medskip

A weighted formulation (for the fuzzy implementation) is possible as well. Assuming that a weight $w_i\in\mathbb{R}$ is associated with every time series $\hat{s}_i$, we can redefine our objective function~\eqref{obj_avg_sax} for the average $r$ as 
\begin{equation}\label{obj_avg_sax_w}
    r =\ \argmin_{s\in \mathcal{S}}\frac{\sum_{i=1}^{m}{w_i\cdot\big(\mindist(\hat{s}_i,s)\big)^2}}{\sum_{i=1}^m{w_i}}
    = \argmin_{s\in \mathcal{S}}\sum_{i=1}^{m}{w_i\cdot\big(\mindist(\hat{s}_i,s)\big)^2}\ .
\end{equation}
Following similar steps to~\eqref{obj_avg_sax}--\eqref{eq:final_final_sax_avg}, we obtain 

 \begin{equation}\label{eq:final_final_sax_avg_w}
    r_j=\argmin_{\beta\in \AAA}{\sum_{i=1}^{m}w_i\cdot\big(\dist(\hat{x}_j^{(i)},\beta)\big)^2}.
\end{equation}
which can be rewritten as

\begin{equation}
\label{eq:final_final_final_sax_avg_w}
    r_j=\argmin_{\beta\in \AAA}{\sum_{\alpha \in \AAA}\Big(n_{\alpha,w}\cdot\big(\dist(\alpha,\beta)\big)^2\Big)}, 
\end{equation}
where $n_{\alpha,w}=\text{sum}\{w_i\text{ } ,i=1,\dots,m$ $|$ $\hat{x}_j^{(i)}=\alpha\}$.

\medskip

Given that Algorithm~\ref{alg:SAX_mean} is based on the Euclidean average, one could suggest that averaging time series before changing the representation into SAX could return the same results.
However, due to the nature of SAX, this is simply not true in a general case.

\end{document}